\documentclass[conference]{IEEEtran}

\usepackage{cite}
\usepackage{amsmath,amssymb,amsfonts}
\usepackage{algorithmic}
\usepackage{graphicx}
\usepackage{rotating}
\usepackage{textcomp}
\usepackage{xcolor}
\usepackage{adjustbox}     
\usepackage{array}
\usepackage{booktabs}      
\usepackage{multirow}     
\usepackage{siunitx}       
\usepackage{caption}
\usepackage{subcaption}    
\usepackage{amsthm} 
\usepackage{times}
\usepackage{epsfig}
\usepackage{hyperref}
\usepackage{calrsfs}
\usepackage{fancyhdr}
\usepackage{comment}
\definecolor{lightmagenta}{RGB}{255,105,180}

\def\BibTeX{{\rm B\kern-.05em{\sc i\kern-.025em b}\kern-.08em
    T\kern-.1667em\lower.7ex\hbox{E}\kern-.125emX}}

\usepackage{calrsfs}
\DeclareMathAlphabet{\pazocal}{OMS}{zplm}{m}{n}

\newcommand{\cmmnt}[1]{}

\begin{document}

\title{2D–3D Feature Fusion via Cross-Modal Latent Synthesis and Attention-Guided Restoration for Industrial Anomaly Detection}

\author{
   \small
   Usman Ali\textsuperscript{1}, Ali Zia\textsuperscript{2}, Abdul Rehman\textsuperscript{1}, 
   Umer Ramzan\textsuperscript{1}, Zohaib Hassan\textsuperscript{1}, Talha Sattar\textsuperscript{1}, Jing Wang\textsuperscript{3}, Wei Xiang\textsuperscript{2} \\
   \textsuperscript{1}GIFT University, Pakistan
   \textsuperscript{2}La Trobe University, Australia
   \textsuperscript{3}Department of Primary Industries, Queensland, Australia\\
   \{usmanali, 211980009, umer.ramzan, 211980050, 211980019\}@gift.edu.pk, \{A.Zia, W.Xiang\}@latrobe.edu.au, jing.wang@dpi.qld.gov.au
}

\maketitle
\begin{abstract}
Industrial anomaly detection (IAD) increasingly benefits from integrating 2D and 3D data, but robust cross-modal fusion remains challenging. We propose a novel unsupervised framework, Multi-Modal Attention-Driven Fusion Restoration (MAFR), which synthesises a unified latent space from RGB images and point clouds using a shared fusion encoder, followed by attention-guided, modality-specific decoders. Anomalies are localised by measuring reconstruction errors between input features and their restored counterparts. Evaluations on the MVTec 3D-AD and Eyecandies benchmarks demonstrate that MAFR achieves state-of-the-art results, with a mean I-AUROC of 0.972 and 0.901, respectively. The framework also exhibits strong performance in few-shot learning settings, and ablation studies confirm the critical roles of the fusion architecture and composite loss. MAFR offers a principled approach for fusing visual and geometric information, advancing the robustness and accuracy of industrial anomaly detection. Code is available at {\color{lightmagenta}\ttfamily\url{https://github.com/adabrh/MAFR}}.
\end{abstract}

\section{Introduction}
Industrial Anomaly Detection (IAD) is a critical component of modern manufacturing, ensuring product quality by identifying rare defects during inspection. These anomalies, which can stem from machine faults, material imperfections, or procedural deviations, must be detected early to mitigate financial losses and maintain high-quality standards \cite{b37}. A fundamental challenge is the inherent scarcity of defective samples, making it difficult to train supervised models. Consequently, the field is dominated by unsupervised methods \cite{b38} that learn a normative model of defect-free data to identify deviations in unseen samples.

Research in IAD has predominantly relied on 2D RGB imagery. However, this approach has significant limitations, as many critical defects are defined by geometry rather than colour. Surface-level anomalies like scratches, dents, or warps can be missed entirely, while inconsistent lighting conditions can obscure visual information or introduce false positives.\cite{b32} To overcome these limitations, the field is increasingly shifting towards multimodal approaches that incorporate 3D surface information\cite{b4}. The fusion of RGB and 3D data provides a comprehensive representation, enabling the robust detection of defects that are subtle or invisible in 2D. This transition has been significantly accelerated by the introduction of benchmark datasets such as MVTec 3D-AD \cite{b4} and Eyecandies \cite{b5}, which provide synchronised multimodal data and have spurred the development of novel methods \cite{b10,b27}.

Recent state-of-the-art models have demonstrated the power of multimodalities. This progress is led by memory-bank-based methods like BTF \cite{b9} and M3DM \cite{b10}, which achieve high performance by comparing test features against a large repository of normal features. For instance, M3DM utilised frozen feature extractors pre-trained on large-scale datasets (ImageNet \cite{imagenet} for RGB, ShapeNet \cite{shapenet} for 3D point clouds) to build a robust feature space. However, this reliance on extensive memory banks can be computationally demanding and memory-intensive, posing challenges for real-time applications. In contrast, teacher-student architectures, such as AST \cite{b27}, offer a more efficient alternative. Despite its speed, AST's performance lags behind the memory-based approaches. This is primarily because it fails to fully exploit the rich geometric information, treating the 3D data as a supplementary channel rather than a distinct structural representation. This suboptimal handling of 3D geometry limits its discriminative power, highlighting a critical gap for a method that can properly leverage 3D structural information. CFM \cite{b32}, excels in accuracy and inference time, but lacks feature fusion across domains.

In this paper, we propose a novel unsupervised framework, Multi-Modal Attention-Driven Fusion Restoration \textit{(MAFR)}, designed to improve 3D-IAD. Our method is based on a feature reconstruction task that leverages both 2D and 3D data. As shown in Figure~\ref{fig:architecture}, the \textit{MAFR} architecture first uses pre-trained feature extractors for the input RGB image and point cloud. The core of our model is a single fusion-encoder that takes both sets of features and combines them into a single, shared latent space. From this unified representation, two separate decoders work in parallel to restore the original 2D and 3D features.  To enhance the accuracy of this process, a Convolutional Block Attention Module (CBAM) is applied in each decoder path to guide the model's focus on the most important features. The entire network is trained end-to-end on anomaly-free samples by optimising a composite loss function. This loss combines the ZNSSD and census losses, which are robust to lighting variations and highly sensitive to local structural patterns, with a smoothness loss that encourages spatially consistent reconstructions. During inference, we measure the discrepancy between the original features and their reconstructed counterparts. Since the model has only learned to accurately restore normal patterns, any anomalous region will result in a high reconstruction error, effectively localising the defect. Our contribution is summarised as: 

\begin{itemize}
    \item We proposed a novel fusion reconstruction model, \textit{MAFR}, that merges 2D and 3D features into a single, unified representation. From this embedding, two parallel decoders reconstruct the original features. This architecture ensures deep, meaningful fusion without the high latency of memory-bank systems.
    
    \item We introduced a composite loss function designed for precise reconstruction. Our training objective combines three distinct components that measure feature similarity, preserve local structural patterns, and enforce surface smoothness. This multi-faceted loss guides the model to learn the characteristics of normal data with high fidelity, making even small anomalies easier to detect.
    
    \item We demonstrated state-of-the-art results on the benchmark datasets  MVTec 3D-AD \cite{b4} and Eyecandies \cite{b5}. Our method outperforms the baseline IAD method. We also showed that it maintains high performance in few-shot learning scenarios, proving its robustness and making it well-suited for industrial applications where training data may be limited.
    
\end{itemize}

\section{Related work}

\subsection{Unsupervised Image Anomaly Detection }\label{AA}
Early work in IAD established two dominant paradigms: reconstruction-based and feature-embedding-based approaches \cite{b11}. The first paradigm centers on image-level reconstruction, using models like autoencoders \cite{b12,b13} or, more recently, diffusion models \cite{b19,b20,multirestore}. The core principle is that a model trained on normal data will fail to accurately reconstruct anomalous regions, with the pixel-level difference between input and output serving as the anomaly score. However, relying on pixel-level discrepancies can be sensitive to non-anomalous variations like lighting and often fails to capture semantic context. To overcome these limitations, a more robust strategy emerged that operates in a learned feature space. Instead of reconstructing the entire image, these methods reconstruct high-level features extracted by a deep neural network \cite{b21}. A significant advancement in this area came with the use of efficient networks (e.g., ResNet, ViT) pre-trained on large-scale datasets \cite{b22,b23,B233,b25,b26,b255}. The rich features from these frozen networks are ideal for memory-bank methods. In this dominant approach, features from normal training samples are stored in a repository. During inference, a test sample's features are compared against this bank, and a large distance to the nearest stored ``normal'' feature signifies an anomaly. The state-of-the-art method in this category, PatchCore \cite{b22}, refines this by creating a memory bank of multi-scale patch-level features and employing coreset subsampling for greater efficiency and performance. Beyond memory banks, other notable paradigms address the problem from different angles. \textit{Distillation-based methods} \cite{b27} train a student network to mimic a teacher on normal data, detecting anomalies where the student's output diverges. Normalizing flows \cite{b28}, conversely, directly model the probability distribution of normal features, identifying low-probability samples as anomalous. 

\subsection{Multimodal Anomaly Detection}\label{Ab}

A common strategy in multi-modal anomaly detection is the use of memory banks to store features from normal samples. A pioneering work, BTF~\cite{b9}, established a foundation by combining pre-trained image features~\cite{pre} with handcrafted 3D features from FPFH~\cite{FPFH} and storing the concatenated results.
Subsequent work aimed to improve upon this foundation in two main ways. First, researchers replaced handcrafted 3D operators with more efficient deep learning models~\cite{DL1,DL2,DL3}. For instance, M3DM~\cite{b10} adopted a pre-trained ViT~\cite{VIT} and PointMAE~\cite{pointmae} for feature extraction, while Shape-Guided~\cite{SHAPE} used a pre-trained PointNet~\cite{DL1}. Second, methods introduced more complex interactions between the two data types. Instead of simple concatenation, M3DM~\cite{b10} built an additional fusion memory bank using contrastive learning, and Shape-Guided~\cite{SHAPE} used 3D features to index and refine the search within the 2D feature bank.
Despite these advances in feature quality and interaction, these methods could not overcome the core limitation of the memory-based approach. The need to build and search through large memory banks, especially multiple ones as in M3DM~\cite{b10}, results in high storage costs and slow inference times. This shared drawback makes it difficult to apply methods like BTF, M3DM, Shape-Guided, and ITNM~\cite{ITNM} in real-time industrial settings.

To overcome the high latency of memory-based methods, an alternative line of research focuses on memory-free architectures. These approaches, however, often introduce a trade-off between inference speed and detection performance.
For example, AST~\cite{b27} uses a teacher-student framework based on knowledge distillation to achieve fast inference. This speed is gained by simplifying the use of 3D data; instead of performing detailed feature extraction, AST treats the depth map as a simple auxiliary channel for concatenation with image features. As a consequence, its ability to localize anomalies is weaker than that of top-performing memory-based models.
Another group of fast methods~\cite{SV1,SV2} relies on anomaly simulation to provide supervised signals for training. While these can be effective, their performance depends heavily on having accurate prior knowledge to generate realistic anomalies, which may limit their use in scenarios with unknown defect types.

A recent method, CFM~\cite{b32}, directly addresses the trade-off between speed and accuracy seen in prior work. Instead of relying on a memory bank, it learns a direct feature mapping between the 2D image and 3D point cloud using a pair of symmetric networks. During inference, a high error in this mapping signals an anomaly.
By using the same backbones as M3DM~\cite{b10}, such as ViT~\cite{VIT} and PointMAE~\cite{pointmae}, but without the memory bank, CFM achieves both high performance and fast inference. However, because the model never explicitly combines the information from both domains into a unified feature representation, it risks losing complementary details. This architectural choice limits its ability to detect anomalies that require the integration of both visual and geometric nature.

\begin{figure*}[htbp]
    \centering
    \includegraphics[width=\linewidth]{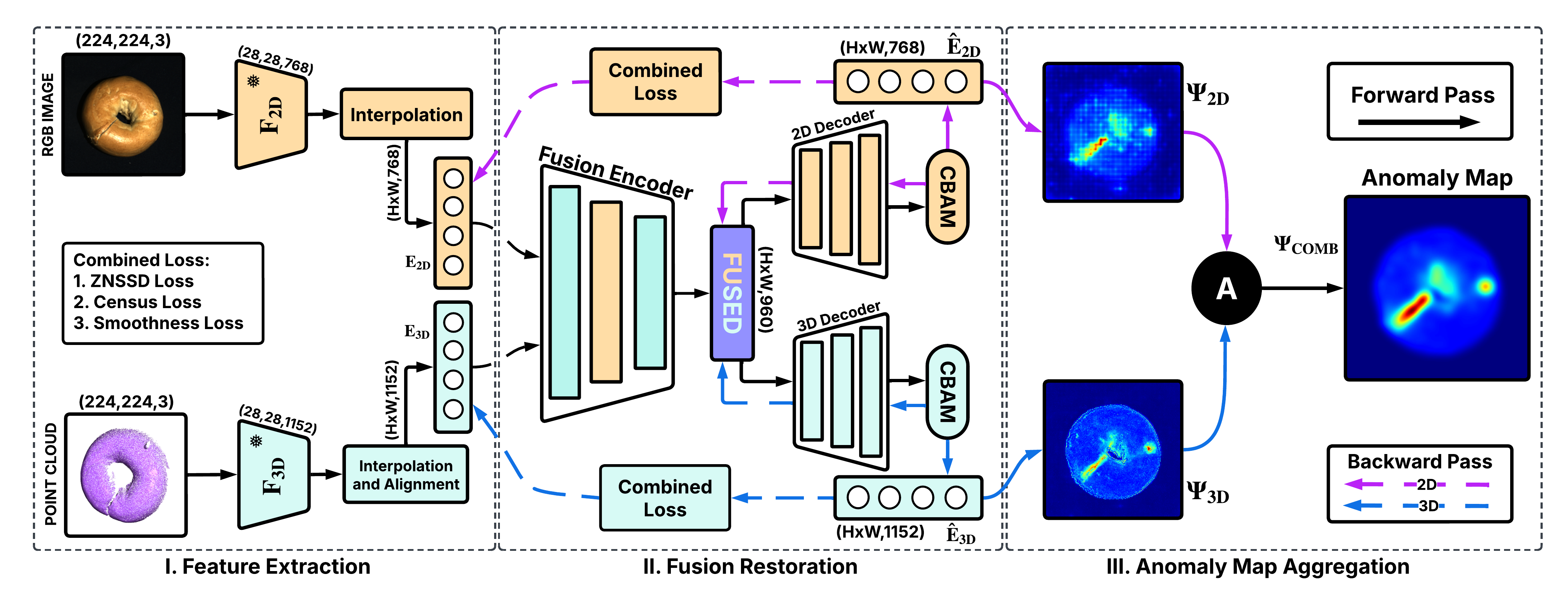}
    \caption{An overview of the proposed architecture. RGB and point cloud inputs are processed by feature extractors ($\mathcal{F}_{2D}$, $\mathcal{F}_{3D}$) to produce deep feature maps. These features follow two paths: (1) A fusion path where an encoder, a fusion block ($F_{\text{fus}}$), and two decoupled decoders generate reconstructed features. (2) A reconstruction path where a combined loss is calculated between the original and reconstructed features. The decoders' outputs, refined by CBAM, produce modality-specific anomaly maps ($\Psi_{2D}$, $\Psi_{3D}$), which are fused via element-wise multiplication (A) to create the final combined anomaly map ($\Psi_{\text{COMB}}$).}
    \label{fig:architecture}
\end{figure*}

\section{Multi-Modal Feature Fusion and Restoration}
Our proposed method, \textit{MAFR}, operates on unsupervised multi-modal fusion reconstruction, identifying defects by detecting inconsistencies between an object's 2D and 3D representations. The framework processes a pair of inputs: an RGB image ($I$) and a corresponding point cloud ($P$). 
The process begins with a feature extraction stage, where pre-trained backbones extract high-level features from both modalities. These features are then passed to the core of our model: a fusion and restoration module. Here, a shared encoder creates a unified latent representation, from which parallel decoders attempt to reconstruct the original features. 
Finally, in the anomaly map generation stage, the reconstruction error between the original and restored features is calculated and combined to pinpoint the location of anomalies.

\subsection{Feature Extraction}
The initial stage of \textit{MAFR} transforms the raw input modalities, an RGB image 
$I \in \mathbb{R}^{H \times W \times 3}$ 
and its corresponding point cloud 
$P \in \mathbb{R}^{N \times 3}$, into spatially-aligned feature representations.This involves a 2D spatial feature  and a 3D geometric feature extraction process to produce the final maps, $E_{\text{2D}}$ and $E_{\text{3D}}$.

\subsubsection{2D Visual Feature Extraction} \label{sec:2d_feat_extraction}
For the RGB image $I$, we used a pre-trained DINO ViT-B/8  \cite{b25} model as our 2D feature extractor, $\mathcal{F}_{\text{2D}}$. DINO ViT-B/8 generates a high-dimensional, low-resolution feature map that captures complex semantic and textural patterns. To restore spatial correspondence with the original input, we applied bilinear interpolation to upsample this map to the original image dimensions. This process yields a dense, pixel-wise feature map 
($E_{\text{2D}} \in \mathbb{R}^{H \times W \times D}$), 
where $D$ is the feature dimension.

\subsubsection{3D Geometric Feature Extraction} \label{sec:3d_feat_extraction}

For the 3D modality, we employed a pre-trained PointMAE \cite{pointmae} model as the feature extractor, $\mathcal{F}_{\text{3D}}$, to generate a feature vector for each point in the input cloud $P$. To enable cross-modal processing, these per-point features are then projected onto the 2D image plane. Since this projection may result in a sparse map with empty pixels (due to occlusions or non-uniform point density), a feature interpolation step is applied to ensure a dense representation. The final output is a 3D-derived feature map, 
$E_{\text{3D}} \in \mathbb{R}^{H \times W \times D}$, 
which is spatially congruent with the image-based feature map $E_{\text{2D}}$. This process effectively translates the 3D geometric information into a 2D spatial format, creating two perfectly aligned feature maps ($E_{\text{2D}}$ and $E_{\text{3D}}$) as input for the subsequent fusion module.

\subsection{Fusion and Restoration}
\label{sec:fusion_restoration}

Following feature extraction, the model employs a fusion and restoration module designed to identify anomalies by learning the normal distribution of multi-modal data. As illustrated in Fig.~\ref{fig:architecture}, this module first uses a fusion encoder to combine the 2D feature map ($E_{\text{2D}}$) and the 3D feature map ($E_{\text{3D}}$) into a unified latent representation ($E_{\text{fus}}$). This shared embedding is then processed by two parallel, decoupled decoders to reconstruct the original features, yielding $\hat{E}_{\text{2D}}$ and $\hat{E}_{\text{3D}}$. The discrepancy between the original and reconstructed features forms the basis for the anomaly score and is minimized during training via a comprehensive loss function.

\subsubsection{Fusion Encoder}
The Fusion Encoder is responsible for creating a shared latent representation that captures complementary information from both imaging and geometric modalities. It takes the aligned feature maps, $E_{\text{2D}}$ and $E_{\text{3D}}$, as input and processes them through a network of transformations (concatenation followed by fully connected layers) to produce a dense, fused embedding, $E_{\text{fus}}$. This process can be expressed as:
\begin{equation}
    E_{\text{fus}} = \phi_{\text{enc}}(E_{\text{2D}}, E_{\text{3D}})
\end{equation}
where $\phi_{\text{enc}}$ represents the learned transformation of the encoder. This unified embedding serves as a robust foundation for the subsequent restoration stage.

\subsubsection{Decoupled Restoration Decoders}
The fused representation $E_{\text{fus}}$ is fed into two independent decoder branches, one for the 2D modality and one for the 3D modality. This decoupled design allows each decoder to specialize in restoring the unique characteristics of its respective data type.

Crucially, each decoder incorporates a convolutional block attention module (CBAM). Rather than being a standalone component, CBAM is integrated directly into the decoding path to refine the features before reconstruction. It sequentially applies channel and spatial attention, enabling each decoder to dynamically focus on the most salient features for its modality. This enhances the model's sensitivity to subtle, localized anomalies while suppressing irrelevant background information.

The output of each decoder is a reconstructed feature map, aiming to be identical to the input for normal samples. The restoration process is defined as:
\begin{equation}
    \hat{E}_{\text{2D}} = \psi_{\text{2D}}(E_{\text{fus}}) \quad \text{and} \quad \hat{E}_{\text{3D}} = \psi_{\text{3D}}(E_{\text{fus}})
\end{equation}
where $\psi_{\text{2D}}$ and $\psi_{\text{3D}}$ are the learnable transformations of the 2D and 3D decoders, respectively, including the CBAM operation.

\subsubsection{Training Objective and Loss Functions}
The entire fusion-restoration network is trained end-to-end by minimizing a combined loss function that measures the reconstruction error between the original features ($E$) and the restored features ($\hat{E}$). This objective is composed of three distinct loss components, ensuring a comprehensive evaluation of similarity, smoothness, and local structure.

\paragraph{Similarity Loss ($L_{\text{sim}}$).} To enforce high fidelity in the reconstruction, we use the Zero-Normalized Sum of Squared Differences (ZNSSD) loss. This metric measures the pixel-wise similarity while being robust to local linear changes in brightness and contrast. It is defined as:
\begin{equation} \label{eq:znssd_loss}
    L_{\text{ZNSSD}}(E, \hat{E}) = \frac{1}{N} \sum_{k=1}^{N} \left( \frac{E_k - \mu_E}{\sigma_E + \epsilon} - \frac{\hat{E}_k - \mu_{\hat{E}}}{\sigma_{\hat{E}} + \epsilon} \right)^2
\end{equation}
The total similarity loss is the sum of the losses for both modalities: $L_{\text{sim}} = L_{\text{ZNSSD}}(E_{\text{2D}}, \hat{E}_{\text{2D}}) + L_{\text{ZNSSD}}(E_{\text{3D}}, \hat{E}_{\text{3D}})$.

\paragraph{Smoothness Loss ($L_{\text{smooth}}$).} This loss imposes an edge-aware regularization penalty on the reconstruction error map ($\Delta E = \hat{E} - E$). It penalizes large gradients in the error map, except where corresponding gradients exist in the original features, thereby preserving sharp details while promoting smoothness elsewhere.
\begin{equation}
\label{eq:smoothness_loss}
\begin{split}
    L_{\text{smooth}}(\Delta E, E) = \frac{1}{N} \sum_{i} \Biggl( &|\nabla_x \Delta E_i|e^{-|\nabla_x E_i|} \\
                                             & + |\nabla_y \Delta E_i|e^{-|\nabla_y E_i|} \Biggr)
\end{split}
\end{equation}
The total smoothness loss is applied to both modalities: $L_{\text{smooth}} = L_{\text{smooth}}(\Delta E_{\text{2D}}, E_{\text{2D}}) + L_{\text{smooth}}(\Delta E_{\text{3D}}, E_{\text{3D}})$.

\paragraph{Census Loss ($L_{\text{census}}$).} To preserve local neighborhood structures and textures, we employ a Census loss. It compares the local spatial patterns around each pixel in the original and reconstructed features using the $L_1$ distance after average pooling, making it robust to outliers.
\begin{equation} \label{eq:census_loss}
    L_{\text{census}}(E, \hat{E}) = \text{Mean}(|\text{AvgPool}(E) - \text{AvgPool}(\hat{E})|)
\end{equation}
The total census loss is: $L_{\text{census}} = L_{\text{census}}(E_{\text{2D}}, \hat{E}_{\text{2D}}) + L_{\text{census}}(E_{\text{3D}}, \hat{E}_{\text{3D}})$.

\paragraph{Total Loss.} The final training objective is a weighted sum of the three components, guiding the model to learn a comprehensive representation of normal data:
\begin{equation} \label{eq:total_loss}
    \mathcal{L}_{\text{total}} = \lambda_1 L_{\text{sim}} + \lambda_2 L_{\text{smooth}} + \lambda_3 L_{\text{census}}
\end{equation}
where $\lambda_1, \lambda_2, \lambda_3$ are hyperparameters that balance the contribution of each loss term. 

\subsection{Anomaly Map Generation}
\label{sec:inference}

During inference, the trained model generates reconstructed feature maps, $\hat{E}_{\text{2D}}$ and $\hat{E}_{\text{3D}}$, from the input features $E_{\text{2D}}$ and $E_{\text{3D}}$. The anomaly score is then derived from the reconstruction error in a multi-step process.

First, we compute modality-specific anomaly maps by calculating the pixel-wise Euclidean distance between the original and reconstructed features:
\begin{equation}
    \Psi_{\text{2D}} = \|E_{\text{2D}} - \hat{E}_{\text{2D}}\|_2, \quad \Psi_{\text{3D}} = \|E_{\text{3D}} - \hat{E}_{\text{3D}}\|_2
\end{equation}
These maps are then fused using element-wise multiplication ($\odot$), which acts as a logical AND operator. This strategy ensures that a high anomaly score is registered only where both modalities jointly indicate a deviation from the norm, significantly reducing false positives.

To handle regions with missing data (e.g., occluded areas in the point cloud), we suppress anomaly scores in pixels corresponding to invalid 3D points. The final combined anomaly map is thus given by:
\begin{equation}
    \Psi_{\text{comb}} = \Psi_{\text{2D}} \odot \Psi_{\text{3D}}
\end{equation}
where scores in invalid regions are subsequently masked to zero.

Finally, for post-processing, the combined map $\Psi_{\text{comb}}$ is smoothed with a Gaussian filter~\cite{b32} to reduce noise and highlight coherent anomalous regions. The resulting smoothed map serves as the final pixel-level anomaly localization result. For sample-level classification, the global anomaly score is taken as the maximum value of this map.

\section{Experimental Settings}

\subsection{Model Architecture}
Our proposed model, \textit{MAFR}, is a multi-modal fusion network built upon a central encoder and two decoupled decoders, where each of these core components is composed of three fully-connected layers. The network takes a $224 \times 224$ pixel RGB image and a corresponding 3D point cloud as input, which are first processed by feature extractors to generate a 768-dimensional 2D feature vector and a 1152-dimensional 3D feature vector. The fusion encoder concatenates these vectors and processes the resulting 1920-dimensional vector to produce a compact 968-dimensional fused embedding. This shared embedding is then fed into the two parallel decoders, which are tasked with reconstructing the original 768-dim 2D and 1152-dim 3D feature spaces. Each intermediate layer within the encoder and decoders is followed by a GELU activation, layer normalization, and a dropout layer (with $p=0.1$) for regularization. To enhance reconstruction fidelity, the decoders are further enhanced with residual skip connections and a CBAM to refine the output features. The entire network is trained jointly for 100 epochs using the adam optimizer with a learning rate of $10^{-3}$, guided by a combined loss function composed of ZNSSD, census, and smoothness losses. All experiments have been conducted on a system with a single \textit{NVIDIA RTX 4080 GPU} featuring \textit{16GB} of dedicated memory.

\subsection{Datasets}
We evaluate our method on two widely used benchmarks for multi-modal anomaly detection: MVTec 3D-AD ~\cite{b4} and Eyecandies~\cite{b5}. The MVTec 3D-AD dataset consists of 10 categories of real-world industrial objects, comprising a total of 2,656 training, 294 validation, and 1,197 test samples. The Eyecandies dataset is a large-scale synthetic benchmark featuring photorealistic food items on an industrial conveyor, providing 10,000 training, 1,000 validation, and 4,000 test samples. A key characteristic of both datasets is the provision of synchronized RGB images and their corresponding 3D point clouds (containing $x,y,z$ coordinates). This ensures precise pixel-level alignment between the appearance (2D) and geometric (3D) modalities, a critical prerequisite for our fusion-based approach.

\subsection{Evaluation Metrics}
We evaluate our model's performance using standard metrics for both benchmarks, addressing both image-level anomaly detection and pixel-level segmentation. At the image level, we report the Image-level AUROC \textit{(I-AUROC)}, which assesses the model's ability to distinguish anomalous samples from normal ones based on a global anomaly score per image. For the more granular task of pixel-wise anomaly segmentation, we employ two complementary metrics. The Pixel-level AUROC \textit{(P-AUROC)} measures overall localization performance by evaluating the classification of all pixels. To provide a more practical measure of localization precision, we also report the area under the per-region overlap curve \textit{(AUPRO)}, which evaluates the quality of detected anomalies by measuring the region-level overlap with the ground-truth masks, making it less sensitive to the large number of correctly classified normal pixels.

\section{Results and Discussion}
\label{sec:results}

This section details the comprehensive evaluation of our proposed method, \textit{MAFR}, establishing its state-of-the-art performance and dissecting the architectural contributions driving its success. We first benchmark \textit{MAFR} against baseline methods on the MVTec 3D-AD and Eyecandies dataset, presenting extensive quantitative results for both per-class and overall performance. To complement these metrics, we provide qualitative visualizations that offer intuitive proof of our model's precise anomaly localization. Furthermore, we scrutinize the model's robustness and data efficiency in challenging few-shot learning scenarios. Finally, we conclude with a series of in-depth ablation studies to validate the critical roles of our proposed loss components and the multi-modal fusion strategy.



\subsection{Comparison with Baseline Methods}
\label{subsec:comparison}

The quantitative performance of \textit{MAFR}, detailed in Table~\ref{tab:mvtec3d} and Table~\ref{tab:eyecandies}, establishes a clear advantage over existing methods on both benchmarks. On the real-world MVTec 3D-AD dataset, our model achieves a mean I-AUROC of 0.972. The per-class results highlight the model's remarkable consistency and versatility, as it delivers highly competitive performance across the full spectrum of object categories. This is particularly evident on classes with challenging geometric defects ('Cable Gland') or subtle textural anomalies ('Tire', 'Carrot'), where our model demonstrates superior discriminative capabilities. This robustness is further validated on the synthetic Eyecandies dataset, where \textit{MAFR} again attains the leading mean I-AUROC of 0.901, underscoring its ability to generalize effectively to different data domains.

Figure~\ref{fig:qualitative}, which provides a qualitative comparison of anomaly maps. Our method consistently generates segmentations with high localization precision and minimal background noise. On challenging samples like 'Cookie' and 'Peach', \textit{MAFR} produces sharp, well-defined heatmaps that precisely contour the ground truth, in stark contrast to the diffuse and often inaccurate activations from competing methods. Furthermore, on objects with strong, repetitive textures like 'Tire' and 'Foam', our model successfully isolates the defect while remaining robust to the object's inherent patterns---a common failure mode for other approaches.

\begin{table*}
\centering 
\caption{Image-level anomaly detection performance (I-AUROC) on the MVTec 3D-AD dataset. The table shows a per-class comparison of our method, \textit{MAFR}, against several state-of-the-art baselines. The mean score across all classes is also reported. The best result in each column is in \textbf{bold}, and the second-best is in \textcolor{blue}{blue}.} 
\label{tab:mvtec3d} 
\small 
\renewcommand{\arraystretch}{1.2}  
\setlength{\tabcolsep}{6pt} 
\begin{tabular}{c||c||cccccccccc||c}
& \textbf{Method} & Bagel & Cable Gland & Carrot & Cookie & Dowel & Foam & Peach & Potato & Rope & Tire & \textbf{Mean} \\
\midrule 
\multirow{12}{*}{\rotatebox[origin=c]{90}{\textbf{I-AUROC}}}
& DepthGAN \cite{b4} & 0.538 & 0.372 & 0.580 & 0.603 & 0.430 & 0.534 & 0.642 & 0.601 & 0.443 & 0.577 & 0.532 \\
& DepthAE \cite{b4} & 0.648 & 0.502 & 0.650 & 0.488 & 0.805 & 0.522 & 0.712 & 0.529 & 0.540 & 0.552 & 0.595 \\
& DepthVM \cite{b4} & 0.513 & 0.551 & 0.477 & 0.581 & 0.617 & 0.716 & 0.450 & 0.421 & 0.598 & 0.623 & 0.555 \\
& VoxelGAN \cite{b4} & 0.680 & 0.324 & 0.565 & 0.399 & 0.497 & 0.482 & 0.566 & 0.579 & 0.601 & 0.482 & 0.517 \\
& VoxelAE \cite{b4} & 0.510 & 0.540 & 0.384 & 0.693 & 0.446 & 0.632 & 0.550 & 0.494 & 0.721 & 0.413 & 0.538 \\
& VoxelVM \cite{b4} & 0.553 & 0.772 & 0.484 & 0.701 & 0.751 & 0.578 & 0.480 & 0.466 & 0.689 & 0.611 & 0.609 \\
& BTF \cite{b9} & 0.918 & 0.748 & 0.967 & 0.883 & 0.932 & 0.582 & 0.896 & 0.912 & 0.921 & 0.886 & 0.865 \\
& AST \cite{b27} & 0.983 & 0.873 & 0.976 & 0.971 & 0.932 & 0.885 & \textcolor{blue}{0.974} & \textbf{0.981} & \textbf{1.000} & 0.797 & 0.937 \\
& M3DM \cite{b10} & \textcolor{blue}{0.994} & \textcolor{blue}{0.909} & 0.972 & 0.976 & 0.960 & \textcolor{blue}{0.942} & 0.973 & 0.899 & 0.972 & 0.850 & 0.945 \\
& CFM \cite{b32} & \textcolor{blue}{0.994} & 0.888 & \textcolor{blue}{0.984} & \textbf{0.993} & \textcolor{blue} {0.980} & 0.888 & 0.941 & 0.943 & 0.980 & \textcolor{blue}
{0.953} & \textcolor{blue}{0.954} \\
& \textit{MAFR} & \textbf{0.997} & \textbf{0.934} & \textbf{0.987} & \textcolor{blue}{0.989} & \textbf{0.983} & \textbf{0.945} & \textbf{0.978} & \textcolor{blue}{0.960} & \textcolor{blue}{0.992} & \textbf{0.963} & \textbf{0.972} \\
\bottomrule
\end{tabular}
\end{table*}

\begin{figure*}[!htbp]
    \centering
    \footnotesize 
    \setlength{\tabcolsep}{1.2pt} 
    \renewcommand{\arraystretch}{1.2} 

    \begin{minipage}[t]{0.49\textwidth}
        \centering
        \begin{adjustbox}{max width=\linewidth}
        \begin{tabular}{@{}c@{}ccc|ccc}
            & \textbf{RGB} & \textbf{PT} & \textbf{GT} & \textbf{M3DM}\cite{b10} & \textbf{CFM}\cite{b32} & \textbf{MAFR} \\ 
            \rotatebox{90}{\textbf{Bagel}} &
            \includegraphics[width=1.3cm]{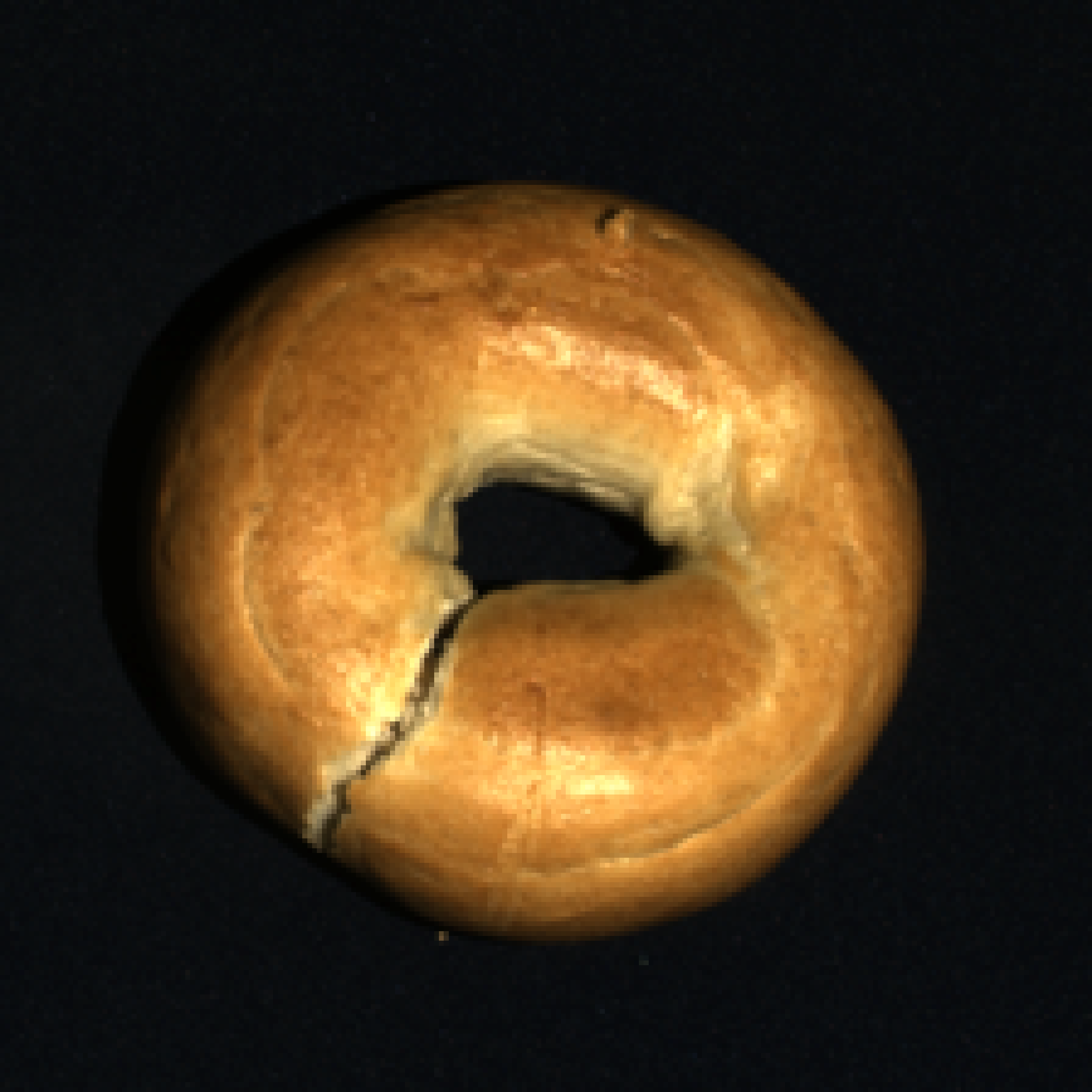} &
            \includegraphics[width=1.3cm]{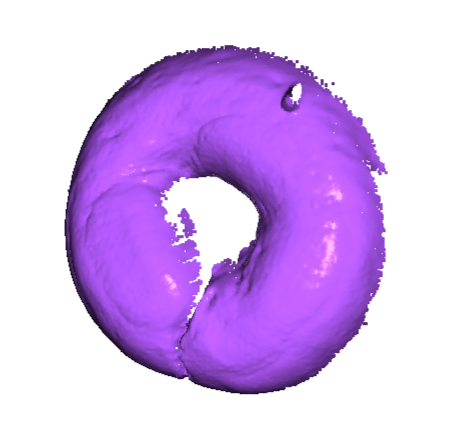} &
            \includegraphics[width=1.3cm]{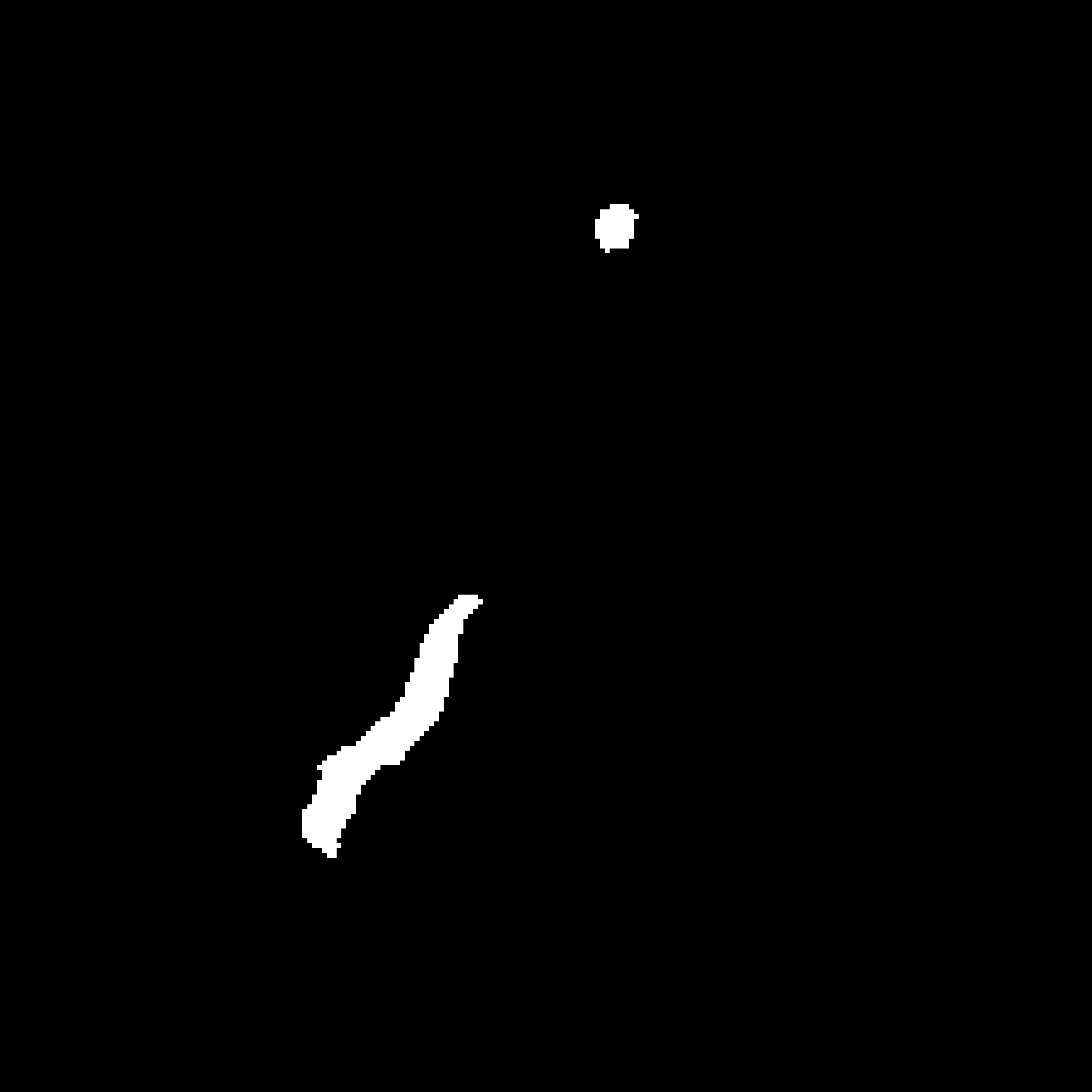} &        \includegraphics[width=1.3cm]{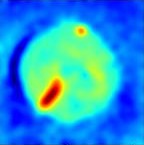} & 
            \includegraphics[width=1.3cm]{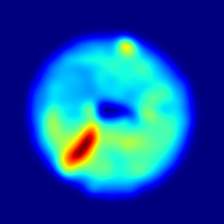} &
            \includegraphics[width=1.3cm]{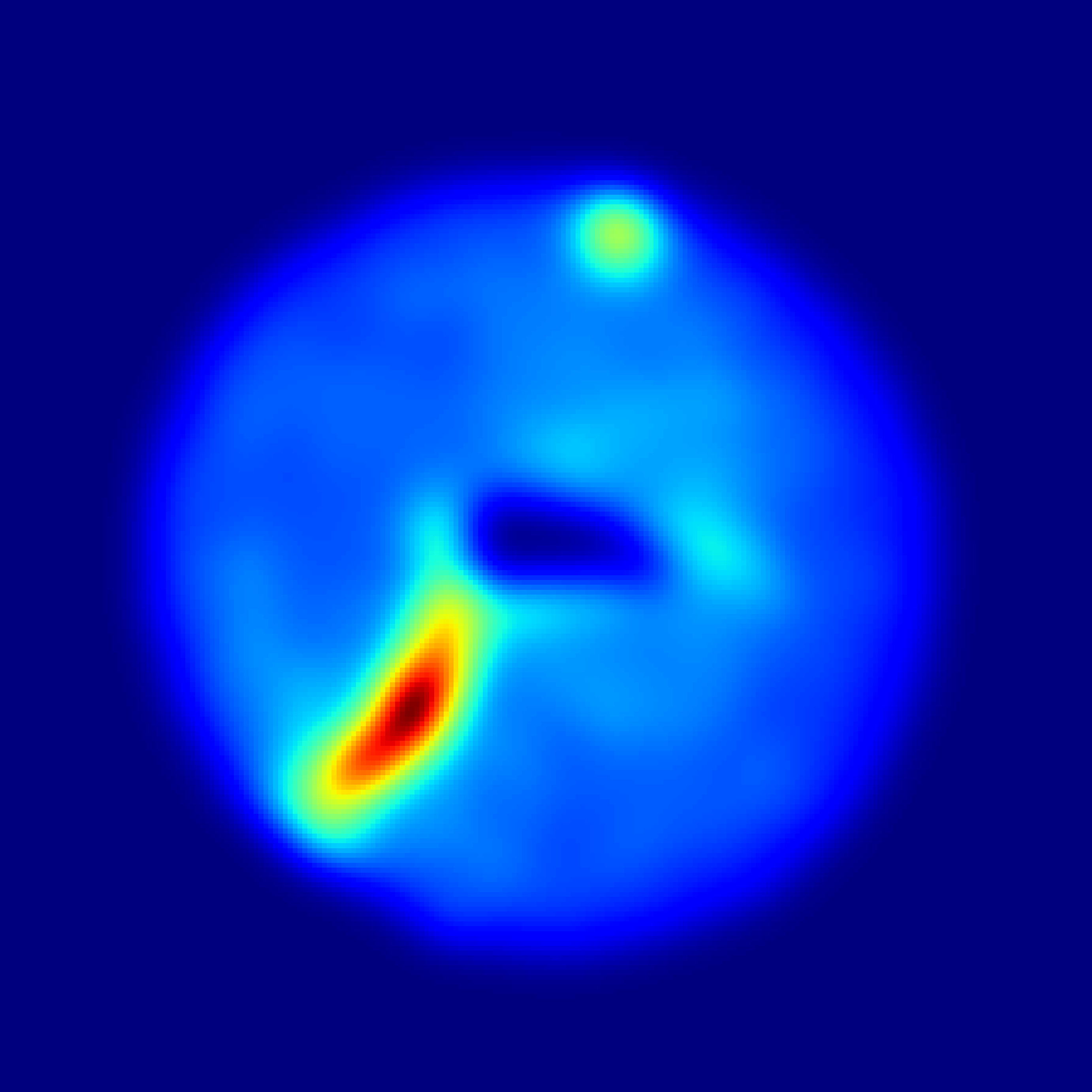}\\ 
    
            \rotatebox{90}{\textbf{Cable G..}} &
            \includegraphics[width=1.3cm]{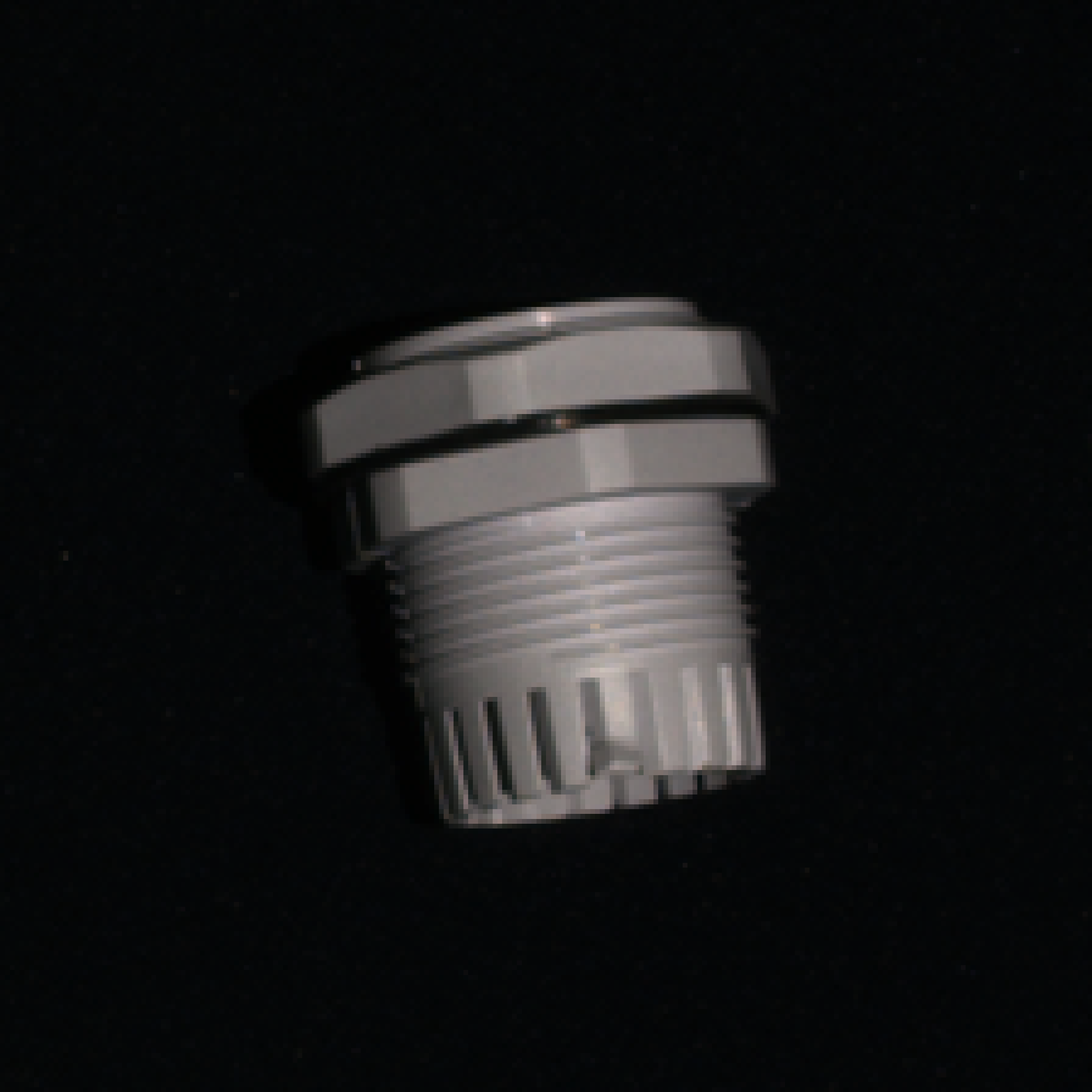} &
            \includegraphics[width=1.3cm]{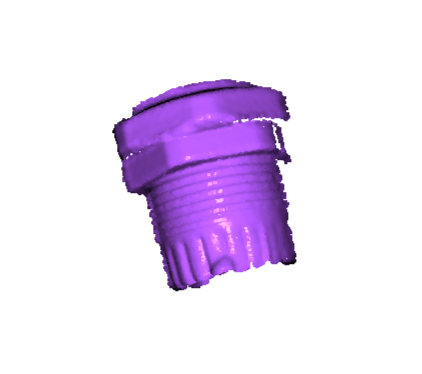} &
            \includegraphics[width=1.3cm]{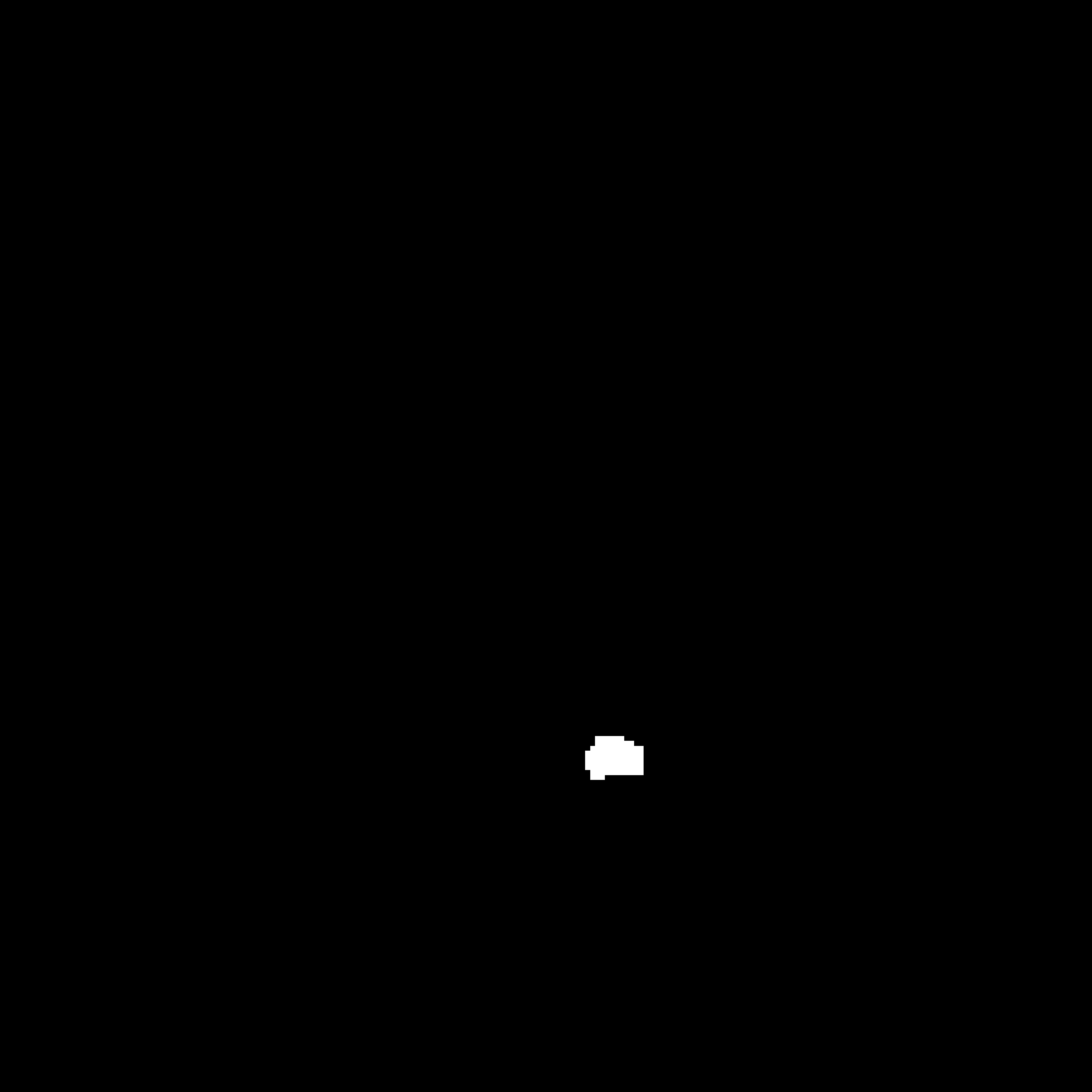} &
            \includegraphics[width=1.3cm]{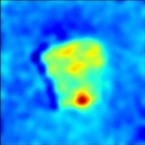} &
            \includegraphics[width=1.3cm]{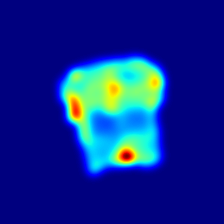} &
            \includegraphics[width=1.3cm]{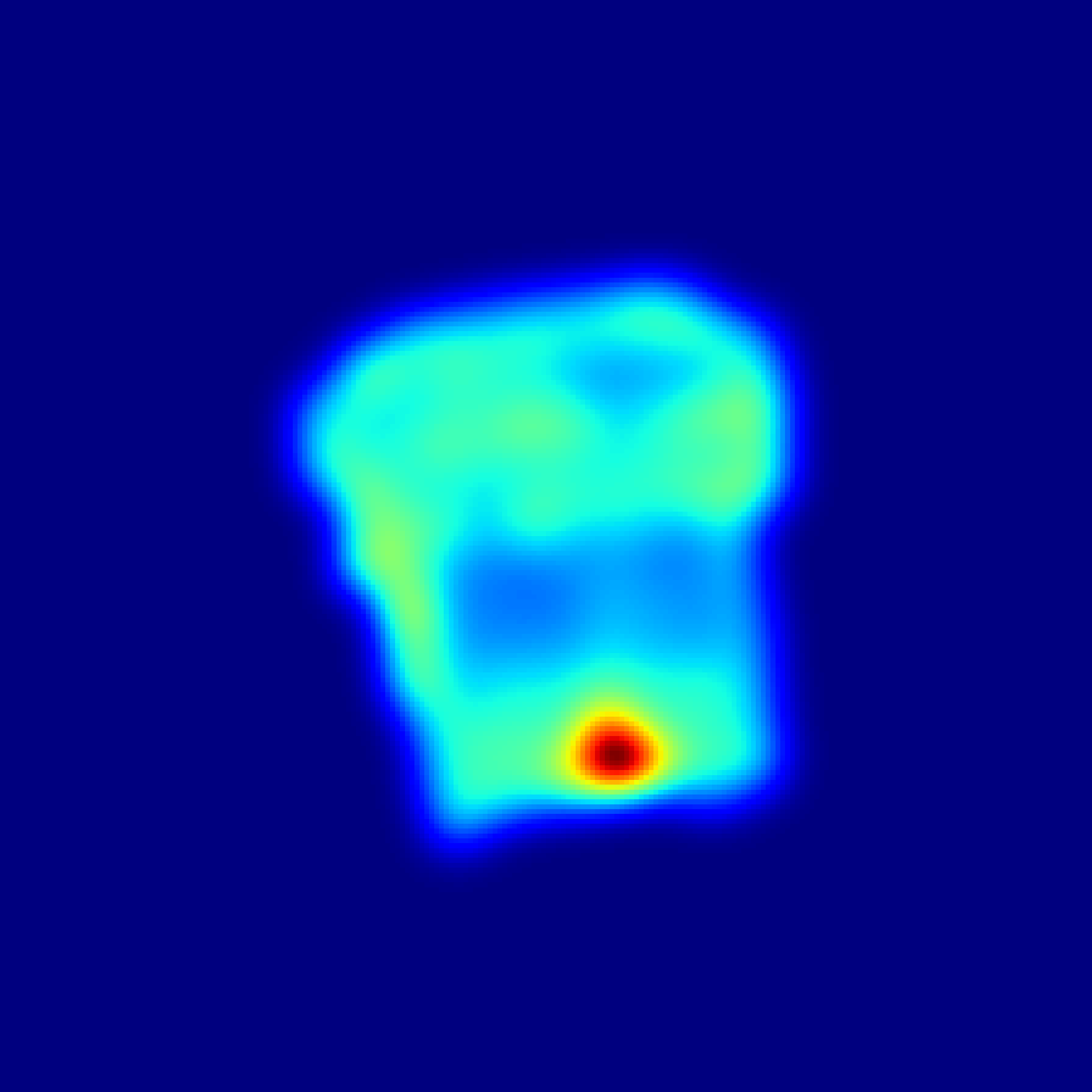}\\ 
    
            \rotatebox{90}{\textbf{Carrot}} &
            \includegraphics[width=1.3cm]{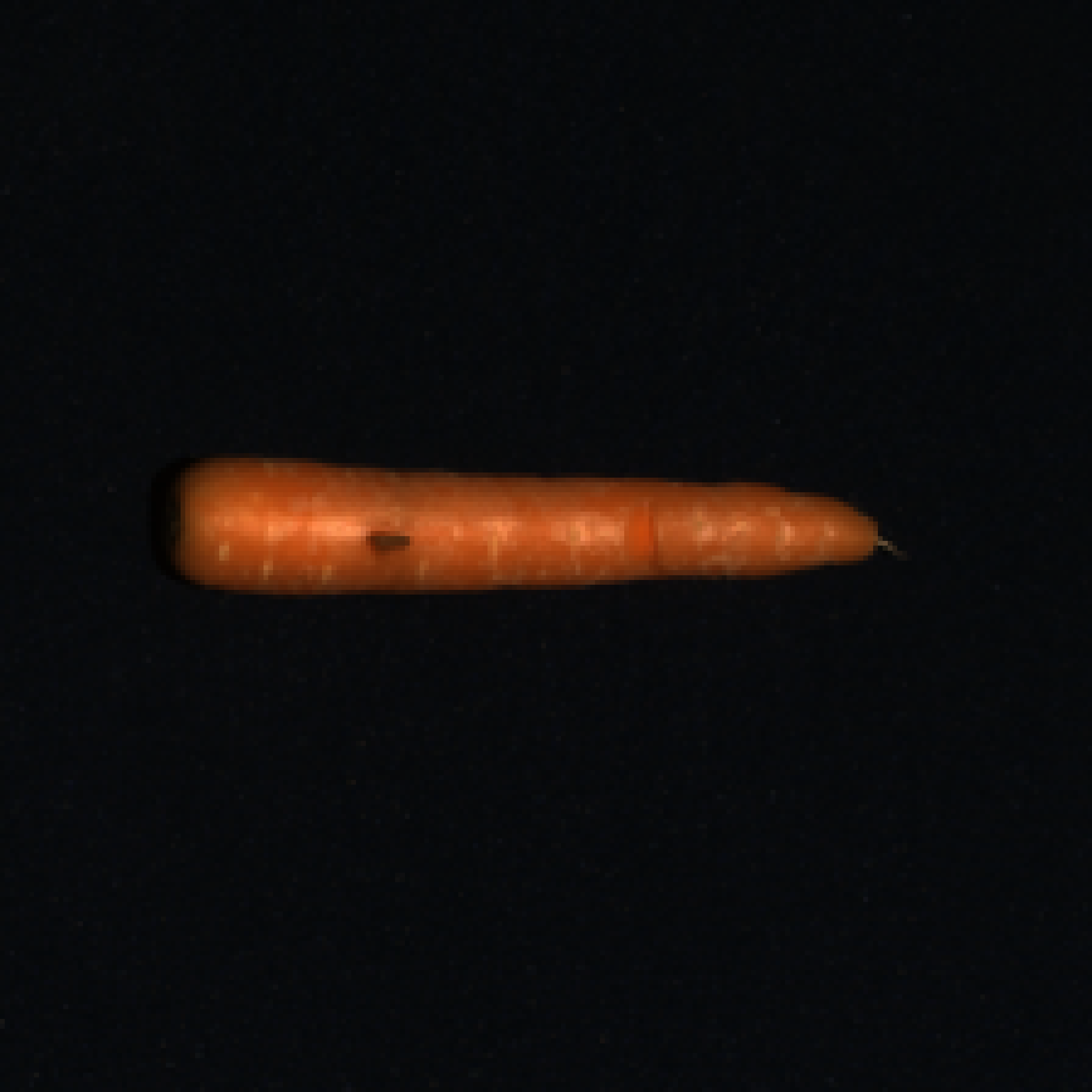} &
            \includegraphics[width=1.3cm]{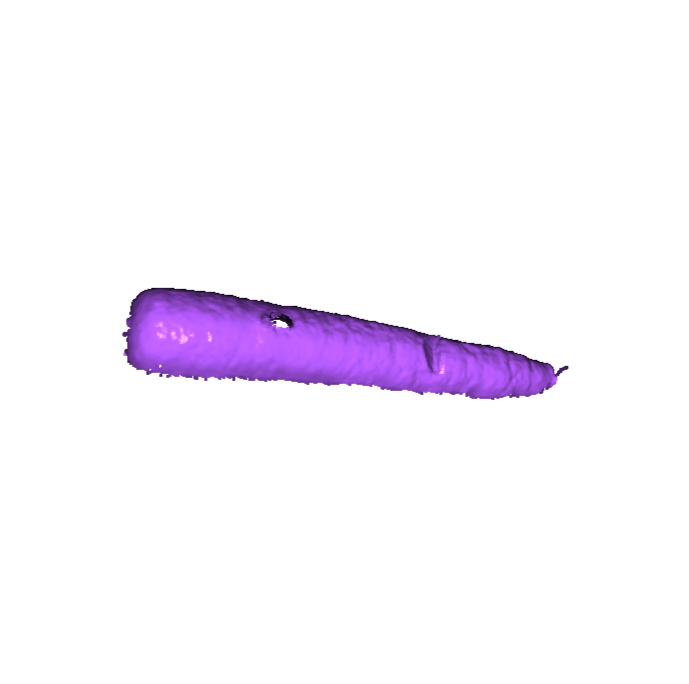} &
            \includegraphics[width=1.3cm]{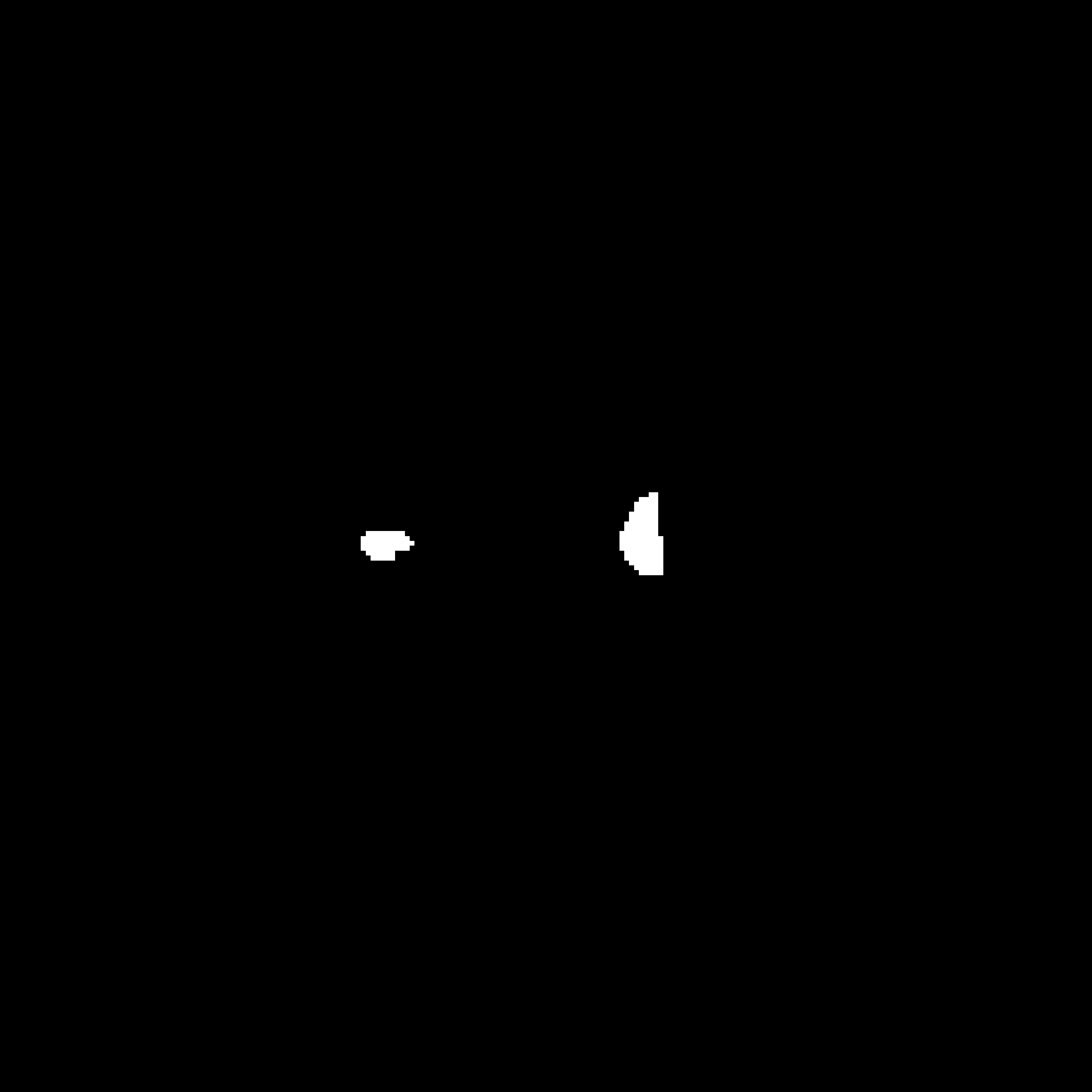} &
            \includegraphics[width=1.3cm]{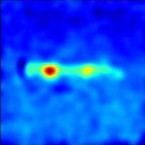} & 
            \includegraphics[width=1.3cm]{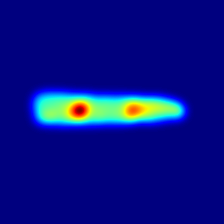} & 
            \includegraphics[width=1.3cm]{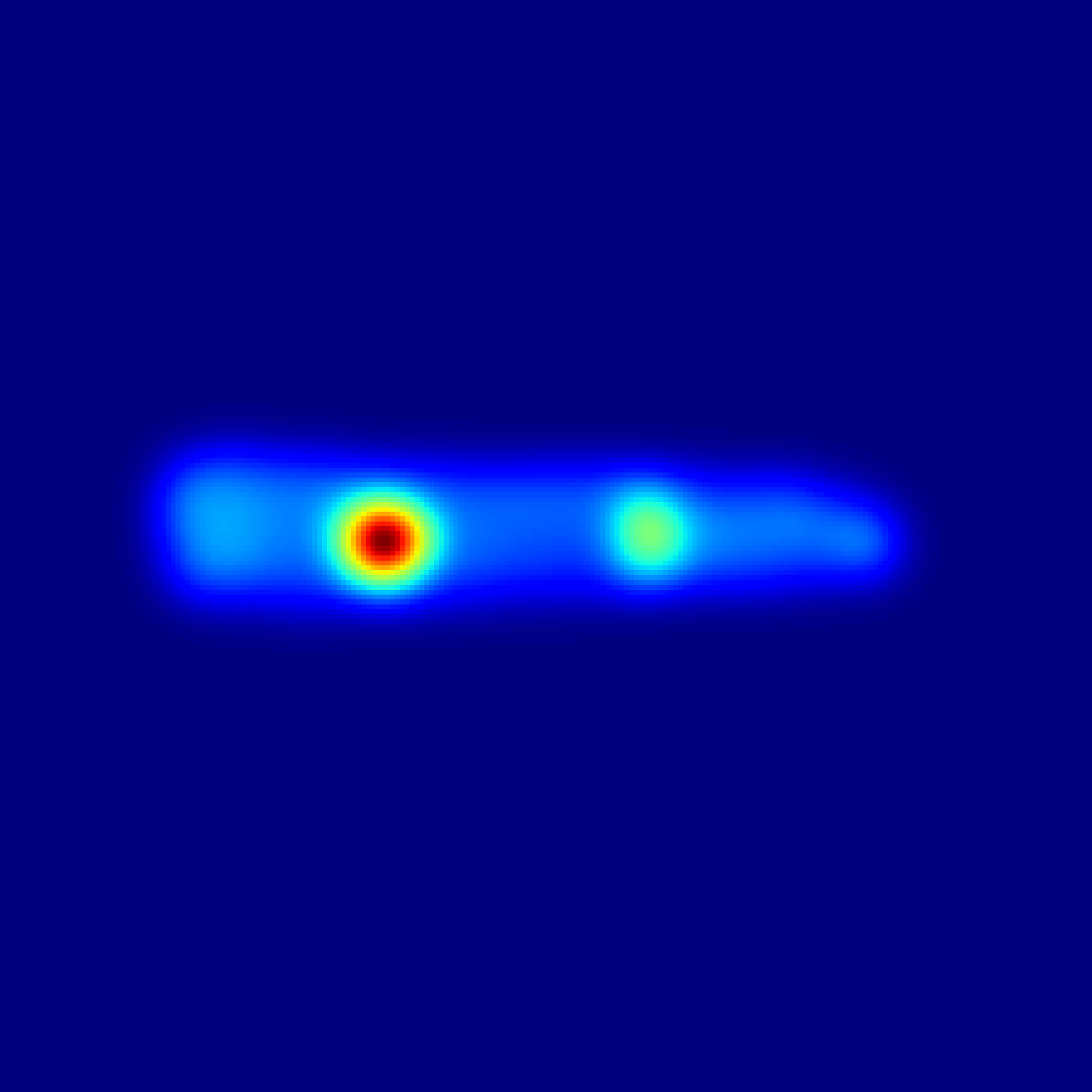}\\ 
    
            \rotatebox{90}{\textbf{Cookie}} & 
            \includegraphics[width=1.3cm]{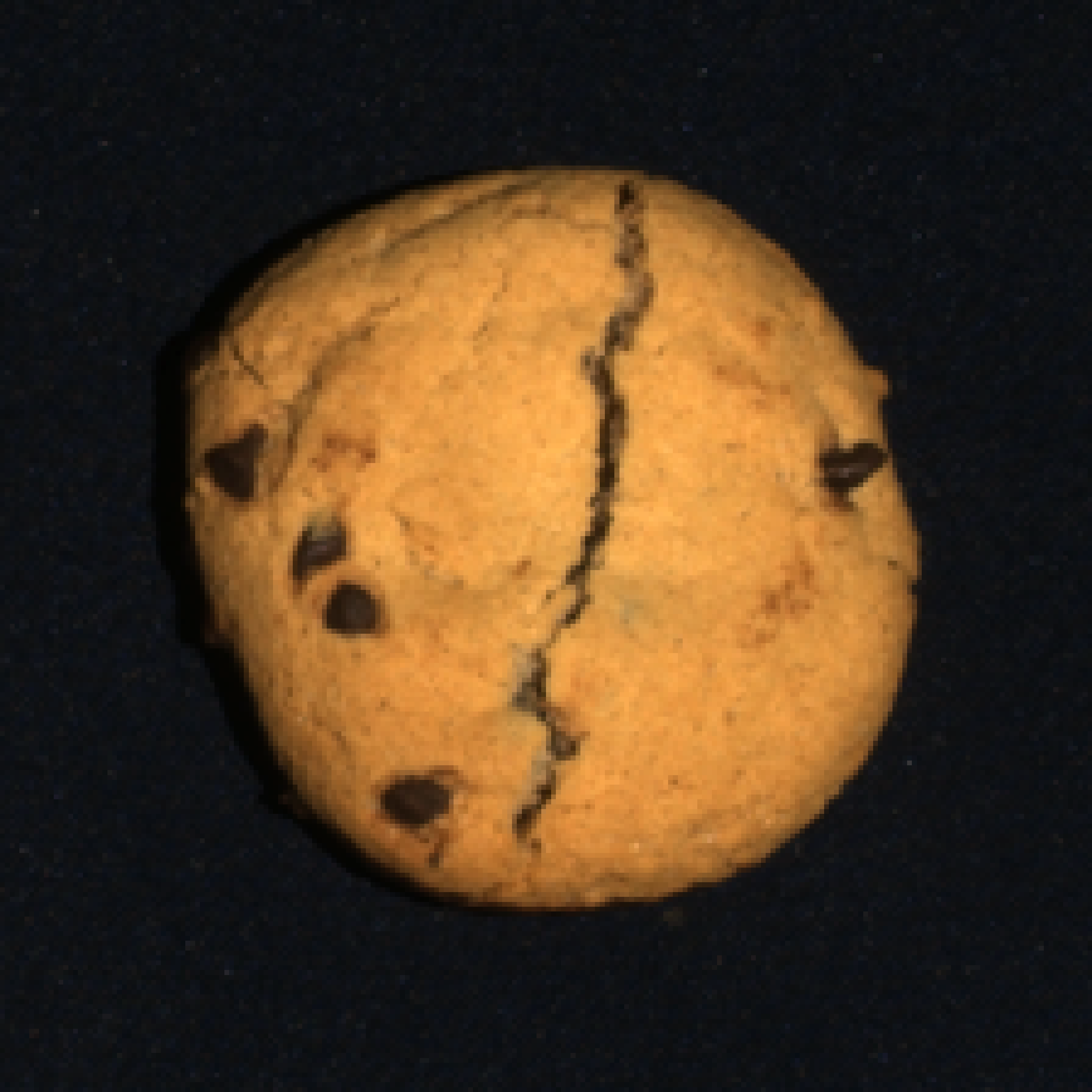} &
            \includegraphics[width=1.3cm]{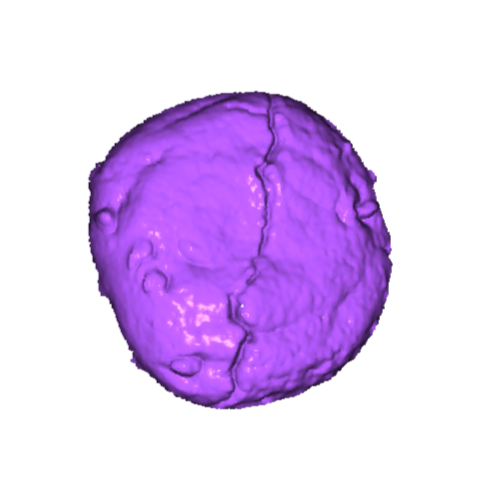} &
            \includegraphics[width=1.3cm]{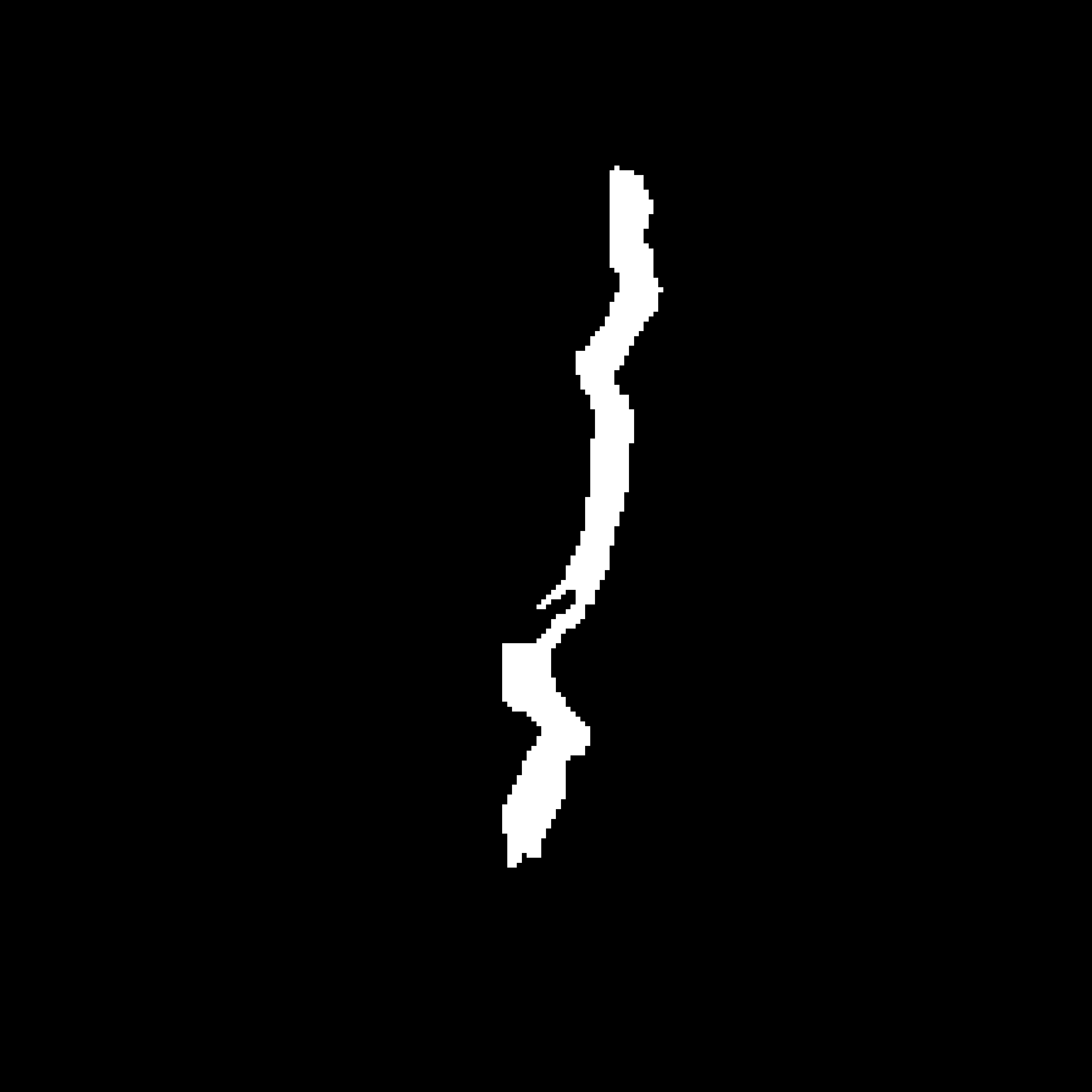} &
            \includegraphics[width=1.3cm]{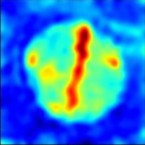} &
            \includegraphics[width=1.3cm]{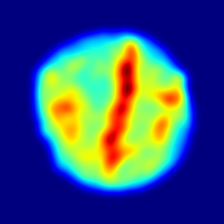} & 
            \includegraphics[width=1.3cm]{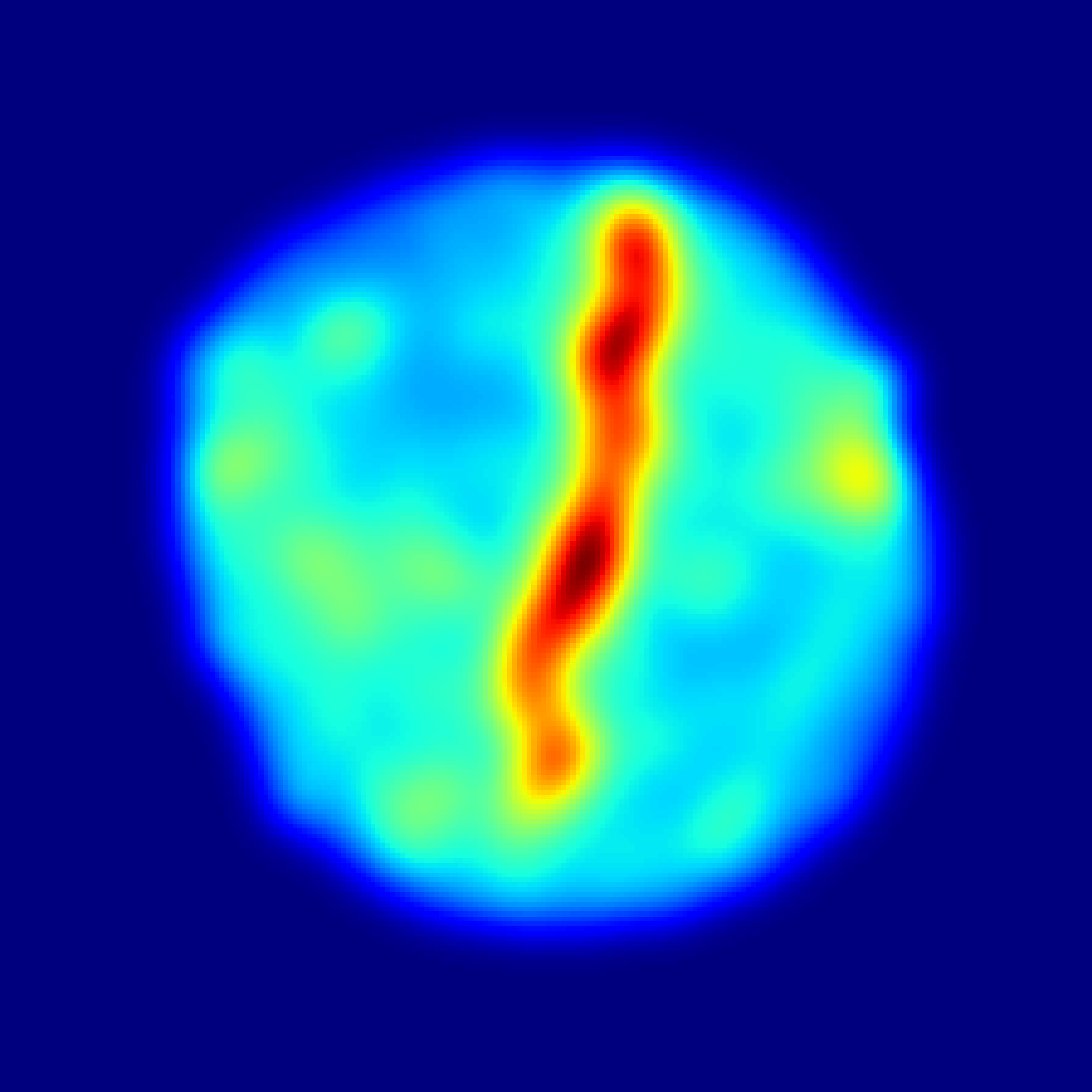}\\
    
            \rotatebox{90}{\textbf{Dowel}} & 
            \includegraphics[width=1.3cm]{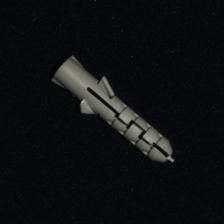} &
            \includegraphics[width=1.3cm]{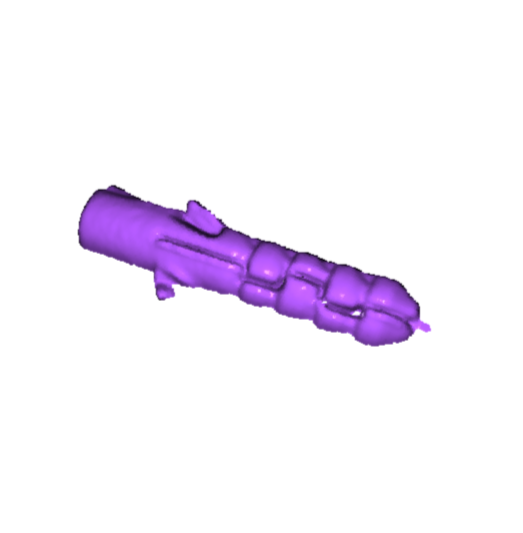} &
            \includegraphics[width=1.3cm]{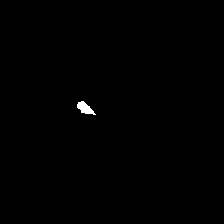} &
            \includegraphics[width=1.3cm]{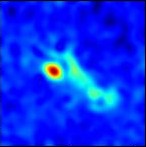} &
            \includegraphics[width=1.3cm]{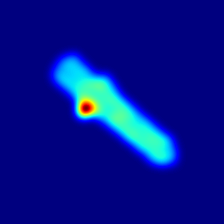} & 
            \includegraphics[width=1.3cm]{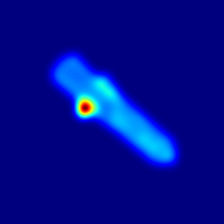}\\
        \end{tabular}
        \end{adjustbox}
    \end{minipage}
    \hfill 
    \begin{minipage}[t]{0.49\textwidth}
        \centering
        \begin{adjustbox}{max width=\linewidth}
        \begin{tabular}{@{}c@{}ccc|ccc}
            & \textbf{RGB} & \textbf{PT} & \textbf{GT} & \textbf{M3DM}\cite{b10} & \textbf{CFM}\cite{b32} & \textbf{MAFR} \\ 
            \rotatebox{90}{\textbf{Foam}} & 
            \includegraphics[width=1.3cm]{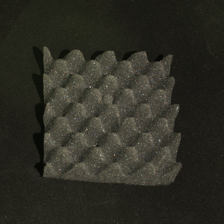} &
            \includegraphics[width=1.3cm]{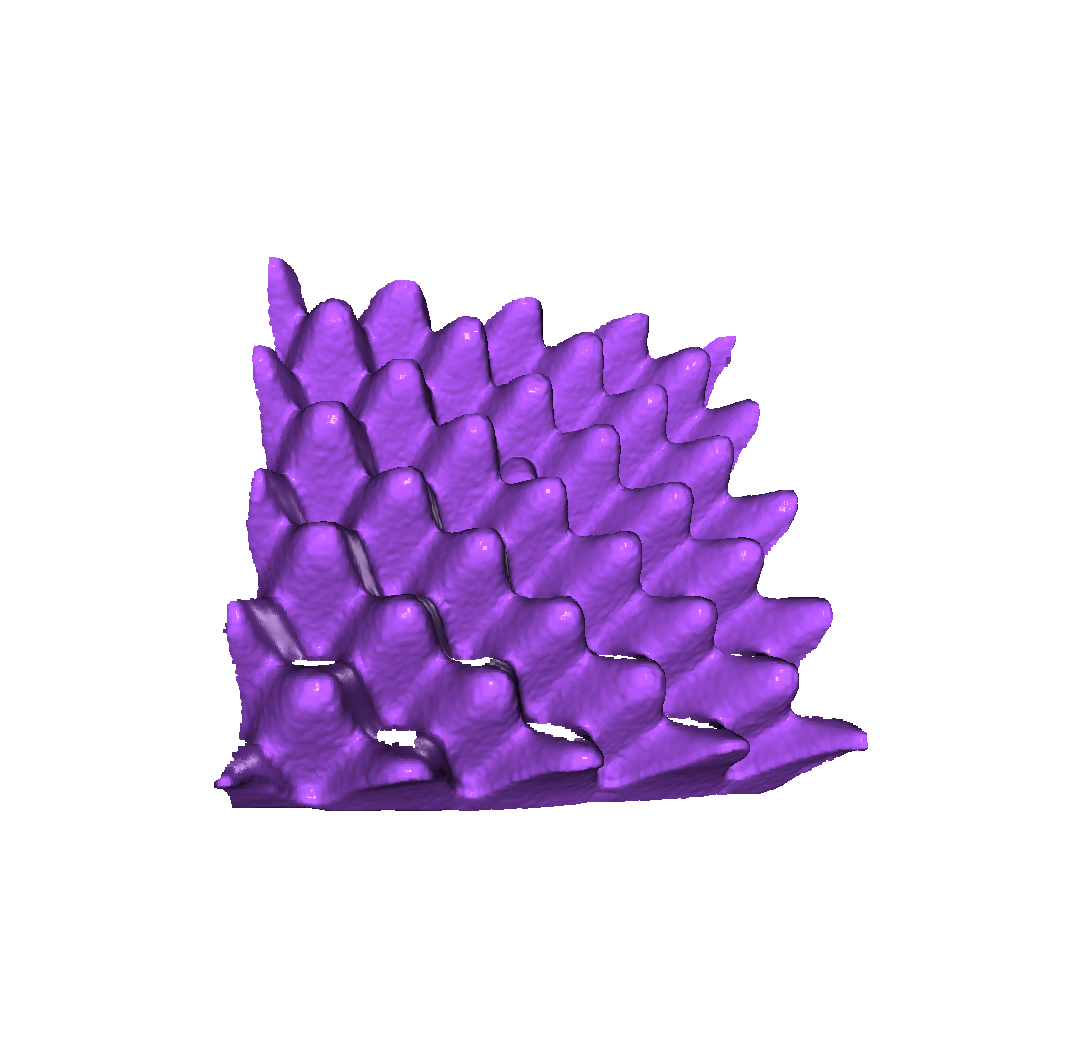} &
            \includegraphics[width=1.3cm]{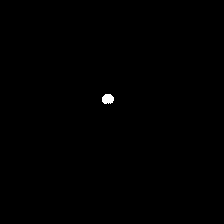} &
            \includegraphics[width=1.3cm]{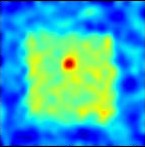} &
            \includegraphics[width=1.3cm]{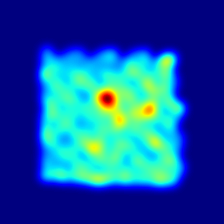} & 
            \includegraphics[width=1.3cm]{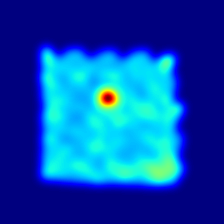}\\
    
            \rotatebox{90}{\textbf{Peach}} & 
            \includegraphics[width=1.3cm]{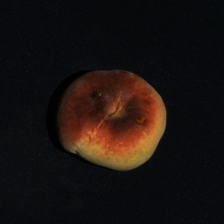} &
            \includegraphics[width=1.3cm]{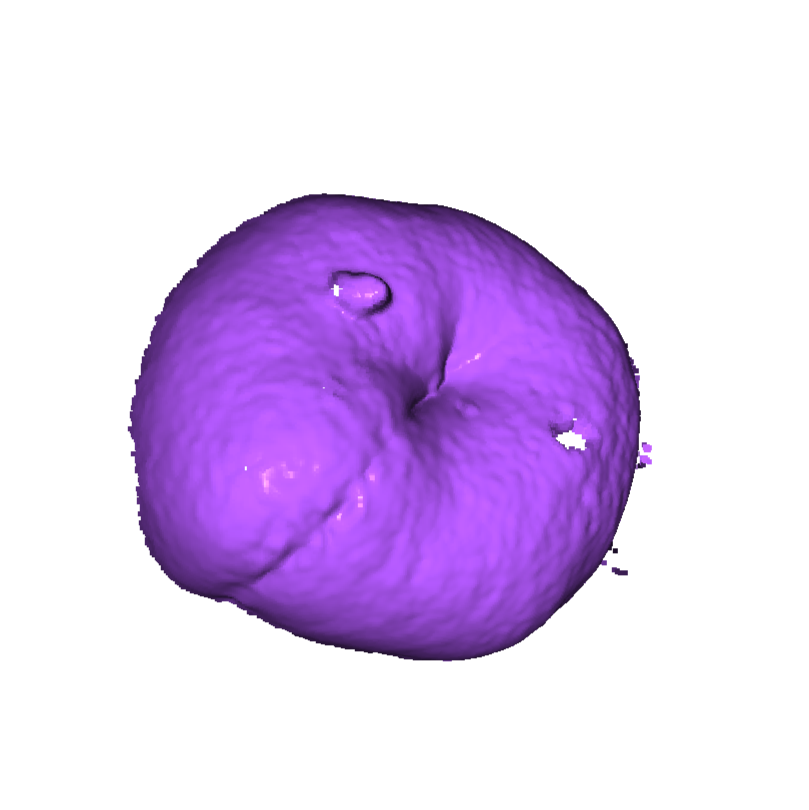} &
            \includegraphics[width=1.3cm]{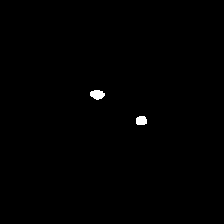} &
            \includegraphics[width=1.3cm]{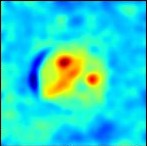} &
            \includegraphics[width=1.3cm]{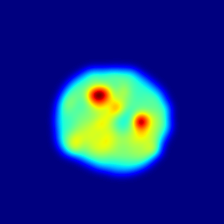} & 
            \includegraphics[width=1.3cm]{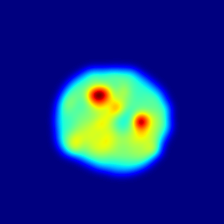}\\
    
            \rotatebox{90}{\textbf{Potato}} & 
            \includegraphics[width=1.3cm]{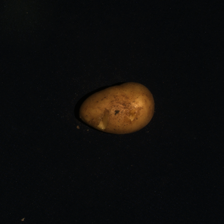} &
            \includegraphics[width=1.3cm]{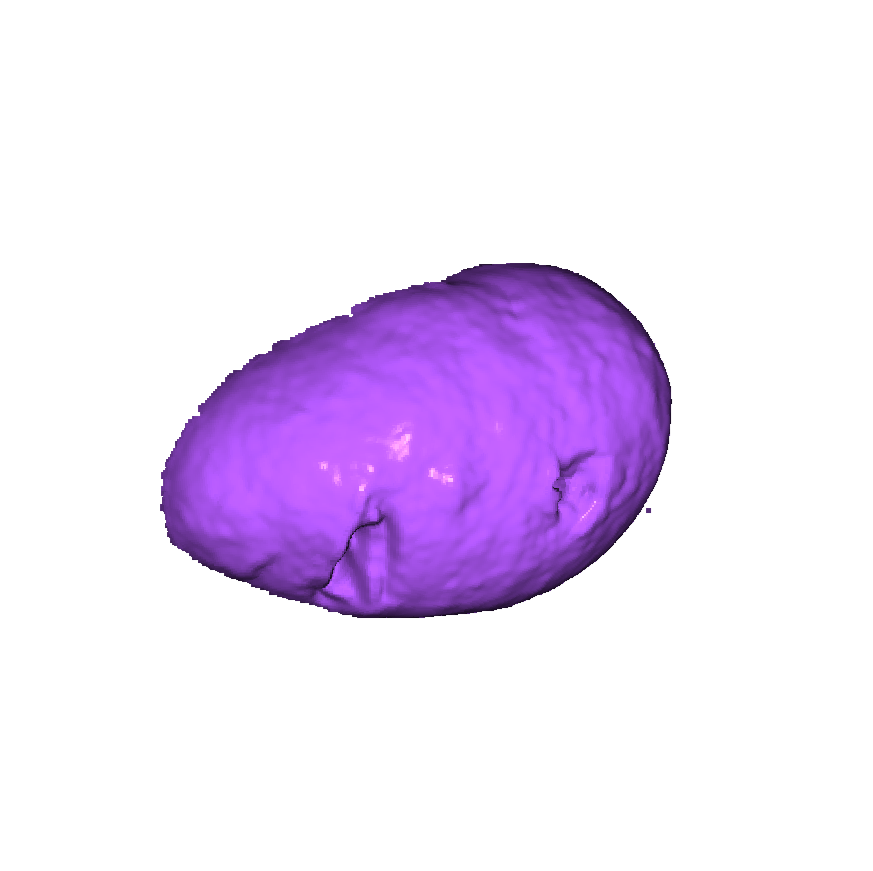} &
            \includegraphics[width=1.3cm]{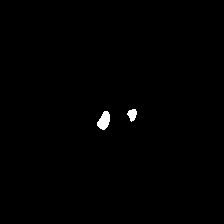} &
            \includegraphics[width=1.3cm]{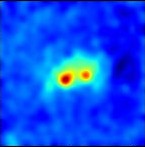} &
            \includegraphics[width=1.3cm]{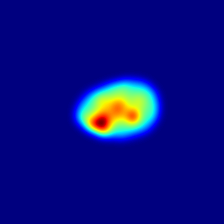} & 
            \includegraphics[width=1.3cm]{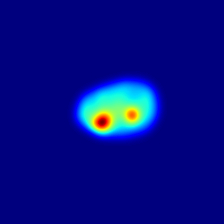}\\ 
    
            \rotatebox{90}{\textbf{Rope}} & 
            \includegraphics[width=1.3cm]{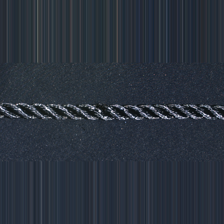} &
            \includegraphics[width=1.3cm]{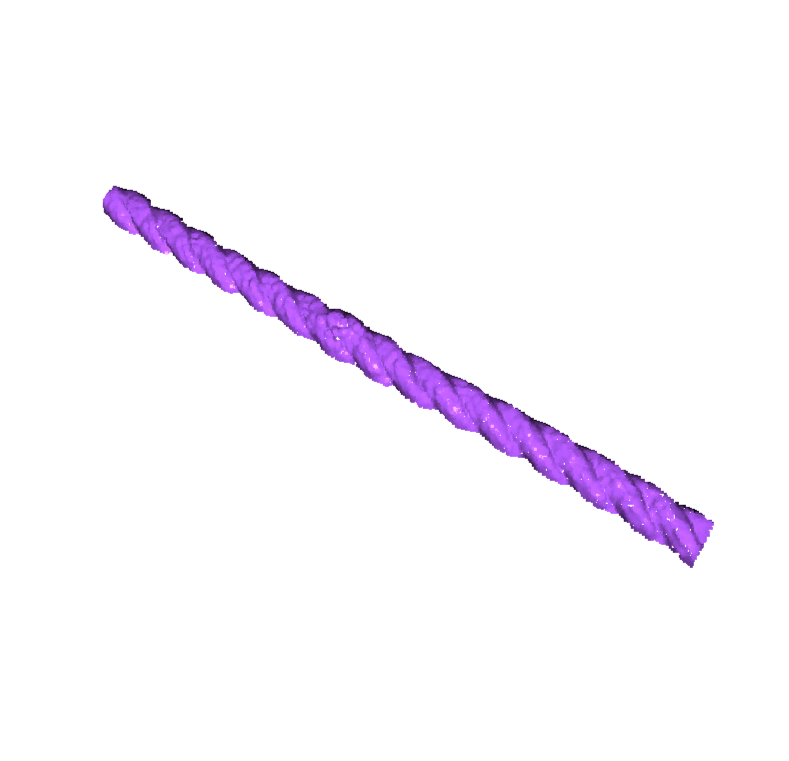} &
            \includegraphics[width=1.3cm]{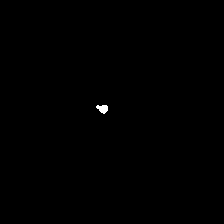} &
            \includegraphics[width=1.3cm]{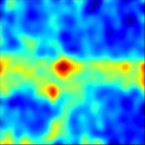} &
            \includegraphics[width=1.3cm]{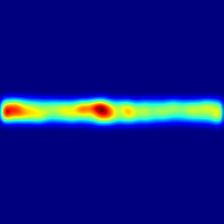} & 
            \includegraphics[width=1.3cm]{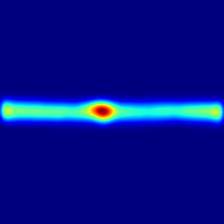}\\ 
    
            \rotatebox{90}{\textbf{Tire}} & 
            \includegraphics[width=1.3cm]{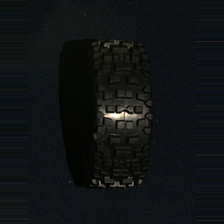} &
            \includegraphics[width=1.3cm]{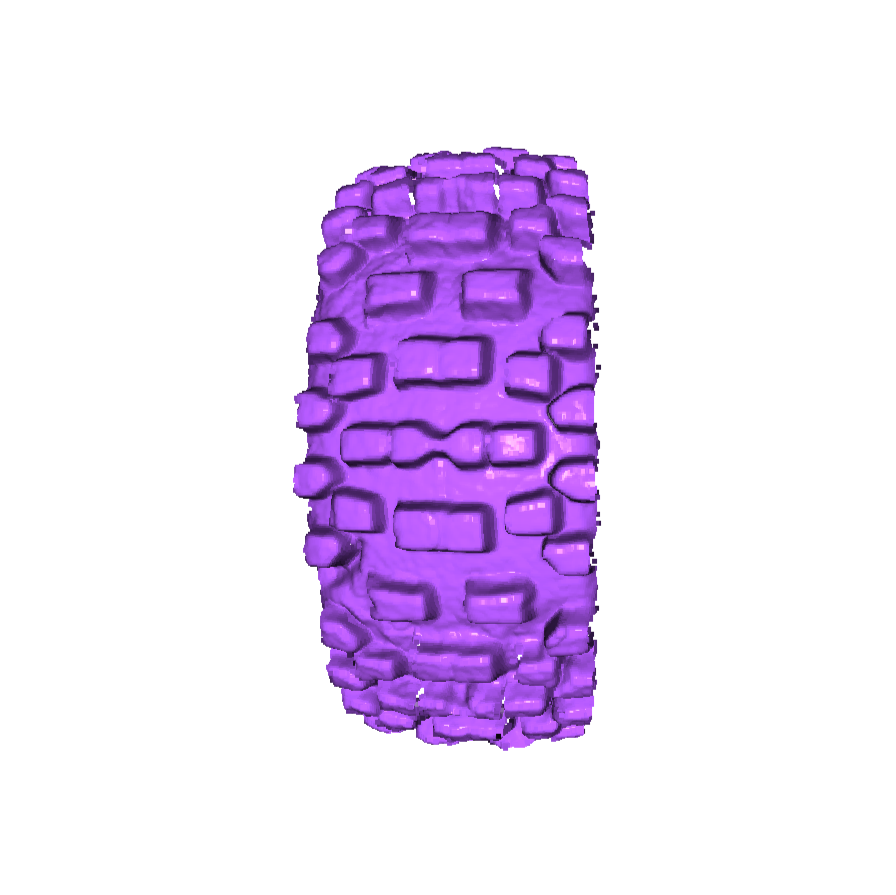} &
            \includegraphics[width=1.3cm]{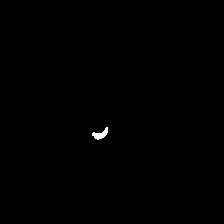} &
            \includegraphics[width=1.3cm]{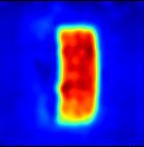} &
            \includegraphics[width=1.3cm]{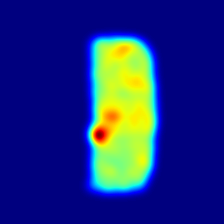} & 
            \includegraphics[width=1.3cm]{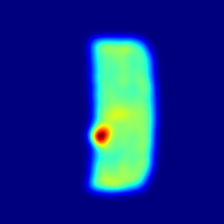}\\ 
        \end{tabular}
        \end{adjustbox}
    \end{minipage}

    \caption{Qualitative results comparison of \textit{MAFR} with the baseline method across 10 categories on MVTec 3D-AD dataset.}
    \label{fig:qualitative}
\end{figure*}

\subsection{Few-Shot Learning Evaluation}

To evaluate the model's performance under data-limited conditions, we conducted a few-shot learning analysis on the MVTec 3D-AD benchmark. This experiment assesses the ability of \textit{MAFR} to function effectively with a reduced number of normal training samples, a frequent constraint in industrial settings. The quantitative results are presented in Table~\ref{tab:my_results}.

For image-level detection (I-AUROC), \textit{MAFR} achieves scores of 0.824 and 0.920 in the 5-shot and 50-shot scenarios, respectively, surpassing M3DM (0.822 and 0.907). On the full dataset, \textit{MAFR} achieves an I-AUROC of 0.972, an improvement of 1.8\%  over CFM.
In pixel-level segmentation (P-AUROC), \textit{MAFR} improves upon the scores of CFM by 0.1 \% and 0.7 \%  in the 5-shot and 50-shot scenarios, respectively. On the full dataset, its score of 0.993 is equivalent to that of CFM.
Regarding the AUPRO@30\% metric, at 5 and 10 shots, \textit{MAFR} records scores of 0.945 and 0.951, while CFM records 0.949 and 0.954, respectively. At 50 shots, however, \textit{MAFR} attains a score of 0.968, which is higher than CFM's 0.965.
Finally, under the AUPRO@1\% metric, the 5-shot score for \textit{MAFR} (0.382) is identical to that of CFM. In the 50-shot and full-dataset settings, its scores of 0.436 and 0.462, respectively, are higher than those of CFM (0.431 and 0.455).


\begin{table*}
\centering 
\caption{ I-AUROC on the Eyecandies dataset. The table shows a per-class comparison of our method, \textit{MAFR}, compared with baselines. The mean score across all classes is also reported. The best result in each column is in \textbf{bold}, and the second-best is in \textcolor{blue}{blue}.} 
\label{tab:eyecandies} 
\small 
\renewcommand{\arraystretch}{1.2}  
\setlength{\tabcolsep}{6pt} 
\begin{tabular}{c||l||cccccccccc||c}
& \textbf{Method} & \textit{Can. C.} & \textit{Cho. C.} & \textit{Cho. P.} & \textit{Conf.} & \textit{Gum. B.} & \textit{Haz. T.} & \textit{Lic. S.} & \textit{Lollip.} & \textit{Marsh.} & \textit{Pep. C.} & \textbf{Mean} \\
\midrule 
\multirow{8}{*}{\rotatebox[origin=c]{90}{\textbf{I-AUROC}}} 
& RGB-D \cite{b5}       & 0.529 & 0.861 & 0.739 & 0.752 & 0.594 & 0.498 & 0.679 & 0.651 & 0.838 & 0.750 & 0.689 \\
& RGB-cD-n \cite{b5}    & 0.596 & 0.843 & 0.819 & 0.846 & 0.833 & 0.550 & 0.750 & 0.846 & 0.940 & 0.848 & 0.787 \\

& AST \cite{b27}        & 0.574 & 0.747 & 0.747 & 0.889 & 0.596 & 0.617 & 0.816 & 0.841 & 0.987 & 0.987 & 0.780 \\
& M3DM \cite{b10}       & 0.624 & \textcolor{blue}{0.958} & \textcolor{blue}{0.958} & \textbf{1.000} & \textcolor{blue}{0.886} & \textcolor{blue}{0.758} & \textcolor{blue}{0.949} & 0.836 & \textbf{1.000} & \textbf{1.000} & \textcolor{blue}{0.897} \\
& CFM  \cite{b32}       & \textcolor{blue}{0.680} & 0.931 & 0.952 & 0.880 & 0.865 & \textbf{0.782} & 0.917 & 0.840 & \textcolor{blue}{0.998} & 0.962 & 0.881 \\
& CFM-M \cite{b32}      & 0.645 & 0.936 & 0.914 & 0.901 & 0.845 & 0.747 & 0.877 & \textbf{0.904} & 0.992 & 0.885 & 0.865 \\
& \textit{MAFR} & \textbf{0.683} & \textbf{0.962} & \textbf{0.960} & \textcolor{blue}{0.995} & \textbf{0.891} & 0.739 & \textbf{0.953} & \textcolor{blue}{0.849} & \textcolor{blue}{0.998} & \textcolor{blue}{0.988}
& \textbf{0.901} \\
\bottomrule
\end{tabular}
\end{table*}

\begin{table*}[h!]
    \centering
    \caption{
    Few-shot performance comparison on the MVTec 3D-AD dataset across multiple metrics and training scenarios.}
    \label{tab:my_results}

    \renewcommand{\arraystretch}{1.1} 
    \setlength{\tabcolsep}{1.1pt}     
    
    \resizebox{\textwidth}{!}{%
    \begin{tabular}{l||cccc||cccc||cccc||cccc}
        & \multicolumn{4}{c||}{\textbf{I-AUROC}} & \multicolumn{4}{c||}{\textbf{P-AUROC}} & \multicolumn{4}{c||}{\textbf{AUPRO@30\%}} & \multicolumn{4}{c}{\textbf{AUPRO@1\%}} \\
        \midrule
        \textbf{Method} & 5-shot & 10-shot & 50-shot & Full & 5-shot & 10-shot & 50-shot & Full & 5-shot & 10-shot & 50-shot & Full & 5-shot & 10-shot & 50-shot & Full \\
        \midrule 
        BTF \cite{b9} & 0.671 & 0.695 & 0.806 & 0.865 & 0.980 & 0.983 & 0.989 & \textcolor{blue}{0.992} & 0.920 & 0.928 & 0.947 & 0.959 & 0.288 & 0.308 & 0.356 & 0.383 \\
        AST \cite{b27} & 0.680 & 0.689 & 0.794 & 0.937 & 0.950 & 0.946 & 0.974 & 0.976 & 0.903 & 0.835 & 0.929 & 0.944 & 0.158 & 0.174 & 0.335 & 0.398 \\
        M3DM \cite{b10} & \textcolor{blue}{0.822} & \textbf{0.845} & \textcolor{blue}{0.907} & 0.945 & 0.984 & \textcolor{blue}{0.986} & 0.989 & \textcolor{blue}{0.992} & 0.937 & 0.943 & 0.955 & 0.964 & \textcolor{blue}{0.330} & 0.355 & 0.387 & 0.394 \\
        CFM \cite{b32} & 0.811 & \textbf{0.845} & 0.906 & \textcolor{blue}{0.954} & \textcolor{blue}{0.986} & \textbf{0.987} & \textcolor{blue}{0.991} & \textbf{0.993} & \textbf{0.949} & \textbf{0.954} & \textcolor{blue}{0.965} & \textbf{0.971} & \textbf{0.382} & \textbf{0.398} & \textcolor{blue}{0.431} & \textcolor{blue}{0.455} \\
        \textit{MAFR} & \textbf{0.824} & \textcolor{blue}{0.841} & \textbf{0.920} & \textbf{0.972} & \textbf{0.987} & 0.983 & \textbf{0.998} & \textbf{0.993} & \textcolor{blue}{0.945} & \textcolor{blue}{0.951} & \textbf{0.968} & \textcolor{blue}{0.970} & \textbf{0.382} & \textcolor{blue}{0.383} & \textbf{0.436} & \textbf{0.462} \\
        \bottomrule
    \end{tabular}%
    }
\end{table*}

\subsection{Ablation Studies}

To validate our design choices and analyze the contribution of each key component, we conducted a series of ablation studies. We investigated the impact of the individual loss functions and the strategy for fusing the final 2D and 3D anomaly maps.

\subsubsection{Analysis of Loss Components}

We first evaluated the effectiveness of each term in our combined loss function. The results, presented in Table~\ref{tab:Loss}, show that the components have distinct and complementary roles. The Census Loss ($L_{\text{census}}$) on its own provides a very strong baseline, achieving an I-AUROC of 0.957 and already excellent segmentation scores. In contrast, the similarity loss ($L_{\text{sim}}$) and smoothness loss ($L_{\text{smooth}}$) are ineffective as standalone objectives, with I-AUROC scores of 0.540 and 0.566, respectively. This indicates they do not learn sufficient discriminative features for anomaly detection on their own.

However, the final row demonstrates their importance as regularizers. The full combination ($L_{\text{sim} + \text{census} + \text{smooth}}$) yields the best overall performance. It improves upon the strong $L_{\text{census}}$ baseline, increasing the I-AUROC from 0.957 to 0.972 and, critically, raising the strict AUPRO@1\% from 0.450 to 0.462. This confirms that while $L_{\text{census}}$ is the primary driver of performance, the inclusion of $L_{\text{sim}}$ and $L_{\text{smooth}}$ provides a significant synergistic benefit, leading to a more robust final model.

\begin{table}[htb]
  \centering
  \setlength{\extrarowheight}{1.5pt}

  \begingroup
  \setlength{\tabcolsep}{1.5pt}

  \caption{Losses Ablation Study}
  \label{tab:Loss}

  \begin{tabular}{l || c c c c} 
      \textbf{Loss Function} & \textbf{I-AUROC} & \textbf{P-AUROC} & \textbf{AUPRO@30\%} & \textbf{AUPRO@1\%}\\ 
      \midrule
      $L_{\text{sim}}$  & 0.540 & 0.850 & 0.529 & 0.006 \\
      $L_{\text{census}}$     & \textcolor{blue}{0.957} & \textbf{0.992} & \textbf{0.968} & \textcolor{blue}{0.450} \\
      $L_{\text{smooth}}$     & 0.566 & \textcolor{blue}{0.891} & \textcolor{blue}{0.640} & 0.033 \\
      $L_{\text{sim} + \text{census} + \text{smooth}}$ & \textbf{0.972}  & \textbf{0.992} & \textbf{0.968} & \textbf{0.462} \\
      \bottomrule
  \end{tabular}
  \endgroup 
\end{table}

\subsubsection{Analysis of Anomaly Map Fusion}

We further analyzed different strategies for combining the 2D anomaly map ($\Psi_{2D}$) and the 3D anomaly map ($\Psi_{3D}$). Table~\ref{tab:anomaly_map_performance} details this comparison. Using only a single modality reveals that the 2D map (I-AUROC of 0.847) is a stronger individual predictor than the 3D map (I-AUROC of 0.802).

Simple fusion strategies like addition ($\Psi_{2D} + \Psi_{3D}$) or taking the maximum value ($\max(\Psi_{2D}, \Psi_{3D})$) already provide a substantial boost over using either modality alone, with addition proving more effective (I-AUROC of 0.920). However, the most effective strategy by a significant margin is element-wise multiplication ($\Psi_{2D} \cdot \Psi_{3D}$). This method improves performance across all metrics, increasing the I-AUROC from 0.920 (for addition) to 0.972. This suggests that multiplication acts as a powerful consensus mechanism, effectively suppressing noise present in only one modality while amplifying confident anomaly predictions that are present in both the 2D and 3D domains. This result strongly validates our choice of multiplicative fusion for the final anomaly map generation.

\begin{table}[htb] 
  \centering 

  \setlength{\extrarowheight}{2pt} 

  \begingroup
  \setlength{\tabcolsep}{2pt}

  \caption{Performance metrics for different anomaly map combinations.} 
  \label{tab:anomaly_map_performance} 

  \begin{tabular}{c || c c c c}
      \textbf{Anomaly Map} & \textbf{I-AUROC} & \textbf{P-AUROC} & \textbf{AUPRO@30\%} & \textbf{AUPRO@1\%}\\ 
      \midrule
      $\Psi_{2D}$ & 0.847 & 0.984 & 0.939 & 0.366 \\
      $\Psi_{3D}$ & 0.802 & 0.967 & 0.899 & 0.309 \\
      $\Psi_{2D} + \Psi_{3D}$ & \textcolor{blue}{0.920} & \textcolor{blue}{0.986} & \textcolor{blue}{0.954} & \textcolor{blue}{0.412} \\
      $\max(\Psi_{2D}, \Psi_{3D})$ & 0.869 & 0.980 & 0.935 & 0.363 \\
      $\Psi_{2D} \cdot \Psi_{3D}$ & \textbf{0.972} & \textbf{0.992} & \textbf{0.968} & \textbf{0.457} \\
      \bottomrule
  \end{tabular}

  \endgroup 
\end{table}

\section{Conclusion}

\textit{MAFR}, a novel approach designed to address the persistent trade-off between accuracy and efficiency in 3D IAD. By unifying 2D and 3D features into a shared latent space and employing attention-guided, decoupled decoders for reconstruction, our method effectively identifies anomalies without computationally expensive memory banks. Our extensive experiments demonstrate that \textit{MAFR} achieves state-of-the-art performance on the MVTec 3D-AD and Eyecandies datasets. The model's robustness was further highlighted in few-shot learning scenarios, where it maintained strong performance even with a severely limited number of training samples, a critical advantage for real-world industrial applications. Furthermore, our ablation studies confirmed that the proposed fusion architecture and composite loss function are essential components driving this success.

Despite these strong results, our work has certain limitations. One notable trade-off is the increased training time associated with the integration of CBAM attention modules within the decoders. While these modules are essential for refining feature reconstruction and boosting localization accuracy, they introduce additional computational overhead during the training phase.

This limitation opens up several promising directions for future research. First, exploring more lightweight attention mechanisms or alternative feature refinement strategies could reduce training costs while preserving performance. Second, a valuable next step would be to extend \textit{MAFR} to an online or continual learning setting, allowing the model to adapt to evolving definitions of "normal" in dynamic industrial environments. Finally, the core principles of our fusion-restoration framework could be adapted to other multimodal challenges, such as integrating thermal or hyperspectral data, further broadening its impact and applicability.

{\small
\bibliographystyle{IEEEtran}
\bibliography{ref}
}
\end{document}